\documentclass[fleqn,10pt]{wlscirep}
\usepackage[utf8]{inputenc}
\usepackage[T1]{fontenc}
\usepackage[font={small,md}]{caption}
\usepackage{subfigmat}
\usepackage{amsmath}
\usepackage{amssymb}
\usepackage{graphicx}
\usepackage{array}
\def\pmb{\boldsymbol}
\usepackage{xcolor}
\usepackage{fontawesome}
\newcolumntype{L}[1]{>{\raggedright\let\newline\\\arraybackslash\hspace{1pt}}m{#1}}
\newcolumntype{C}[1]{>{\centering\let\newline\\\arraybackslash\hspace{1pt}}m{#1}}

\title{Data segmentation based on the local intrinsic dimension} 

\author[1,2]{Michele Allegra}
\author[2]{Elena Facco}
\author[3]{Francesco Denti}
\author[2,4,*]{Alessandro Laio}
\author[5,6,+]{Antonietta Mira}

\affil[1]{Institut de Neurosciences de la Timone UMR 7289, Aix Marseille Universit\'e, CNRS, 13385 Marseille, France}
\affil[2]{Scuola Internazionale Superiore di Studi Avanzati, Trieste, Italy}
\affil[3]{University of California, Irvine, California, U.S.A.}
\affil[4]{International Centre for Theoretical Physics, Trieste, Italy}
\affil[5]{Universit\`a della Svizzera italiana, Lugano, Switzwerland}
\affil[6]{Universit\`a dell'Insubria, Como, Italy}

\affil[*]{laio@sissa.it}
\affil[+]{antonietta.mira@usi.ch}

\begin{abstract}

One of the founding paradigms of machine learning is that a small number of variables is often sufficient to describe high-dimensional data. The minimum number of variables required is called the intrinsic dimension (ID) of the data. Contrary to common intuition, there are cases where the ID varies within the same data set. This fact has been highlighted in technical discussions, but seldom exploited to analyze large data sets and obtain insight into their structure. 
Here we develop a robust approach to  discriminate regions with different local IDs and segment the points accordingly. Our approach is computationally efficient and can be proficiently used even on large data sets.
We find that many real-world data sets contain regions with widely heterogeneous dimensions. These regions host points differing in core properties: folded vs unfolded configurations in a protein molecular dynamics trajectory, active vs non-active regions in brain imaging data, and firms with different financial risk in company balance sheets. A simple topological feature, the local ID, is thus sufficient to achieve an unsupervised segmentation of high-dimensional data, complementary to the one given by clustering algorithms. 
\end{abstract}

\begin{document} 

\flushbottom
\maketitle

\thispagestyle{empty}

\section*{Introduction}

From string theory to science fiction, the idea that we might be “glued” onto a low-dimensional surface embedded in a space of large dimensionality has tickled the speculations of scientists and writers alike. 
However, when it comes to multidimensional data such a situation is quite common rather than wild speculation: data  often concentrate on hypersurfaces of low \textit{intrinsic dimension} (ID).
 Estimating the ID of a dataset is a routine task in machine learning: it yields important information on the global structure of a dataset, and is a necessary preliminary step in several analysis pipelines. 
 
In most approaches for dimensionality reduction and manifold learning, the ID is assumed to be constant in the dataset. 
 This assumption is implicit in projection-based  estimators, such as 
 Principal Component Analysis (PCA) and its variants~\cite{jolliffe1986principal}, Locally Linear Embedding~\cite{roweis2000nonlinear}, and Isomap~\cite{tenenbaum2000global}; and it also underlies geometric ID estimators~\cite{grassberger2004measuring,levina2005maximum,rozza2012novel}, which  infer the ID from the distribution of between-point distances. 
The hypothesis of a constant ID complies with simplicity and intuition but is not necessarily valid. 
In fact, many authors have considered the possibility of ID variations  within a dataset~\cite{barbara2000using,gionis2005dimension,costa2005estimating,carter2010local,campadelli2013local,johnsson2015low,mordohai2005unsupervised,haro2008translated,souvenir2005manifold,wang2010multi,goh2007segmenting,vidal2011subspace,elhamifar2011sparse,elhamifar2013sparse,amsaleg2015estimating,faranda2017dynamical}, often proposing to classify the data according to  this feature. However, the dominant opinion in the community is still that a variable ID is a peculiarity, or a technical detail, rather than a common feature to take into account before performing a data analysis.
This perception is  in part due to the lack of sufficiently general 
methods to track ID variations.
Many of the methods developed in this field make assumptions that limit their applicability to specific classes of data.
Refs.~\cite{barbara2000using,campadelli2013local,johnsson2015low,mordohai2005unsupervised,haro2008translated} use local ID estimators that implicitly or explicitly assume a uniform density. Refs.~\cite{gionis2005dimension,costa2005estimating,carter2010local} jointly estimate the density and the ID from the scaling of neighbor distances, by approaches which work well only if the density varies slowly and is approximately constant in a large neighborhood of each point. Ref.~\cite{souvenir2005manifold} requires a priori knowledge on the number of manifolds and their IDs. 
Refs.~\cite{wang2010multi,goh2007segmenting,vidal2011subspace,elhamifar2013sparse} all require that the manifolds on which the data lay are hyperplanes, or topologically isomorphic to hyperplanes.  These assumptions (locally constant density, and linearity in a suitable set of coordinates) are often violated in the case of real-world data. Moreover, many of the above approaches~\cite{barbara2000using,campadelli2013local,johnsson2015low,mordohai2005unsupervised,souvenir2005manifold,wang2010multi,goh2007segmenting,vidal2011subspace,elhamifar2011sparse,elhamifar2013sparse} work explicitly with the coordinates of the data, while in many applications only  distances between pairs of data points are available.
To our knowledge, only refs.~\cite{amsaleg2015estimating,faranda2017dynamical} do not make any assumption about the density, as they derive a parametric form of the distance distribution using extreme-value theory, which in principle is valid independently of the form of the underlying density. However, they assume that the low tail of the distance distribution is well approximated by its asymptotic form, an equally non-trivial assumption. 

In this work, we propose a method to segment data based on the local ID  which aims at overcoming the aforementioned limitations.  
Building on TWO-NN~\cite{facco2017estimating}, a recently proposed ID estimator which is insensitive to density variations and uses only the distances between points, 
we develop  a Bayesian framework that allows identifying, by Gibbs sampling, the regions in the data landscape in which the ID can be considered constant. 
Our approach works even if the data are embedded on highly curved and twisted manifolds, topologically complex and not isomorphic to hyperplanes, and if they are harvested from a non-uniform probability density. Moreover, it is specifically designed to use only the distance between data points, and not their coordinates. 
These features, as we will show, make our approach computationally efficient and more robust than other methods.
Applying our approach to  data of various origins, we  show that ID variations between different regions are common. These variations often reveal fundamental properties of the data: for example, unfolded states in a molecular dynamics trajectory of a protein lie on a manifold of a lower dimension than the one hosting the folded states.
Identifying regions of different dimensionality in a dataset 
can thus be a way to perform  an unsupervised segmentation of the data.
At odds with common approaches, we do not group the data according to their density but perform segmentation based on a geometric property defined on a local scale: the number of 
independent directions along which neighboring data are spread.

\section*{Methods}

\subsubsection*{A Bayesian approach for discriminating manifolds with different IDs}

The starting point of our approach is the recently proposed TWO-NN estimator~\cite{facco2017estimating}, which infers the IDs from the statistics of the distances of the first two neighbors of each point.
Let the data  $(x_1, x_2,...,x_N)$, with $N$ indicating the number of points, be i) independent and identically distributed samples from a probability density $\rho(x)$ with support on a manifold with unknown dimension $d$, such that ii) for each point $x_i$, $\rho(x_i)$ is approximately constant in the spherical neighborhood with center $i$ and radius given by the distance between
$i$ and its second neighbor. Assumption i) is shared  all methods that infer the ID from the distribution of distances (e.g., Refs. \cite{levina2005maximum,rozza2012novel,costa2005estimating,carter2010local,campadelli2013local,johnsson2015low,haro2008translated}. Assumption ii) is the key assumption of TWO-NN: while of course it may not be
perfectly satisfied, other methods used to infer the ID from distances require uniformity of the density on a larger scale including the first $k$ neighbors, usually with $k\gg2$.  
Let $r_{i1}$ and $r_{i2}$ 
be the distances of the first and second neighbor of $x_i$, then  $ \mu_i \doteq \frac{r_{i2}}{r_{i1}}$ follows the Pareto distribution 
$f(\mu_i | d)=d \mu_i^{-(d+1)}$. 
This readily allows the estimation of $d$ from ${\mu_i}, i=1,2,...,N$. We can write the global likelihood of $\pmb{\mu} \equiv \{ \mu_i \}$ as 
\begin{equation}
P(\pmb{\mu}|d)=d^N\prod_{i=1}^N \mu_i^{-(d+1)}=d^N e^{-(d+1)V},
\label{basic}
\end{equation}
where $V \doteq \sum_{i=1}^N \log(\mu_{i})$. From Eq. (\ref{basic}), and upon specifying a suitable prior on $d$, a Bayesian estimate of $d$ is immediately obtained. This estimate is particularly robust to density variations: in~\cite{facco2017estimating} it was shown that restricting the uniformity requirement only to the region containing the first two neighbors allows properly estimating the ID also in cases where $\rho$ is strongly varying. \\
TWO-NN can be extended to yield a heterogeneous-dimensional model with an arbitrarily high number of components. Let  $\boldsymbol{x}=\{ x_i \}$ be i.i.d samples from a density $\rho(x)$ with support on the union of $K$ manifolds with varying dimensions.
This multi-manifold framework is common with many  previous works investigating heterogeneous dimension in a dataset~\cite{gionis2005dimension,costa2005estimating,souvenir2005manifold,carter2010local,haro2008translated,campadelli2013local,xiao2010data,goldberg2009multi,wang2010multi,elhamifar2011sparse,elhamifar2013sparse,vidal2011subspace}.
Formally, let $\rho(x) = \sum_{k=1}^K p_k \rho_k(x) $ where each $\rho_k(x)$ has support on a manifold of dimension $d_k$ and 
$\pmb{p} \doteq (p_1, p_2,...,p_K)$
are the a priori probabilities that a point belongs to the manifolds $1,\dots,K$. We shall first assume that $K$ is known, and later show how it can be estimated from the data. The distribution of the $\mu_i$ is simply a mixture of Pareto distributions:
\begin{equation}
P(\mu_i | \pmb{d}, \pmb{p} ) \doteq \sum_{k=1}^K p_k d_k \mu_i^{-(d_k+1)}.
\end{equation}
Here, like in ref.~\cite{facco2017estimating}, we assume that the variables $\mu_i$ are independent. This condition could be rigorously imposed by restricting the product in eq. \ref{basic} to data points with no common first and second neighbors. In fact, one does not find significant differences between the empirical distribution of $\pmb{\mu}$ obtained with all points and the one obtained by restricting the sample to the independent points (i.e., the empirical distribution of $\mu$, obtained considering entire sample, is well fitted by the theoretical Pareto law). However, restricting the estimation to independent samples generally worsens the estimates, as one is left with a small fraction (typically, $\approx 15\%$) of the original points. Therefore, for the purpose of parametric inference, we decide to include all points in the estimation. In SI (Table~S1) we show that restricting the estimation to independent points yields results that are qualitatively in good agreement with those obtained using all points, but with generally worse dimension estimates. Following the customary approach~\cite{richardson1997bayesian} we introduce latent variables $ \pmb{z} \doteq (z_1, z_2,...,z_K)$ where $z_i = k$ indicates that 
point $i$ belongs to manifold $k$. We have  $ P(\mu_i | \pmb{d}, \pmb{p}, \pmb{z}  ) = P(\mu_i | z_i, \pmb{d}) P_{pr}(z_i |\pmb{p}) $ with
with
$P(\mu_i | z_i, \pmb{d} )= d_{z_i} \mu_i^{-(d_{z_i}+1)}$, $P_{pr}(z_i |\pmb{p})=p_{z_i}$. 
This yields the posterior 
\begin{equation}
\begin{split}
P_{post}(\pmb{z},\pmb{d},\pmb{p} | \pmb{\mu})
\propto  P (\pmb{\mu}| \pmb{z},\pmb{d}) P_{pr}(\pmb{z}|\pmb{p}) P_{pr}(\pmb{d}) P_{pr}(\pmb{p}).
\end{split}
\label{Eq: model1}
\end{equation}
We use independent Gamma priors on $ \pmb{d}$, $d_k \sim Gamma(a_k,b_k)$ and a joint Dirichlet prior on $ \pmb{p} \sim Dir(c_1,\dots,c_K) $. 
We fix $a_k, b_k, c_k =1 \ \forall k $, corresponding to a maximally non-informative prior on the $\pmb{p}$ and an expectation of  $\mathbb{E}\left[d_k\right]=1$ for all $k$. If one has a different prior expectation on $\pmb{d}$ and $\pmb{p}$, other choices of prior may be more convenient.

The posterior \eqref{Eq: model1}
does not have an analytically simple form, but it can be sampled by standard Gibbs sampling~\cite{casella1992explaining}, allowing for the joint estimation of $\pmb{d},\pmb{p},\pmb{z}$.  
However, model \eqref{Eq: model1} has a serious limitation: Pareto distributions with (even largely) different values of $d$ overlap to a great extent. Therefore, the method can not be expected to correctly estimate the $z_i$: 
a given value $\mu_i$ may be compatible with several manifold memberships. This issue can be addressed by correcting an unrealistic feature of model \eqref{Eq: model1}, namely, 
the independence of the $z_i$.
We assume that the neighborhood of a point is more likely to contain points from the same manifold than from different manifolds. This requirement can be enforced with an additional term in the likelihood that penalizes local inhomogeneity (see also Ref.~\cite{haro2008translated}).
Consider the $q$-\emph{neighbor matrix} $\boldsymbol{\mathcal{N}}^{(q)}$ with nonzero entries $\mathcal{N}_{ij}^{(q)}$ only if 
$j \neq i $ is among the first $\ensuremath{q}$ neighbors of $i$.
Let $\xi$ be the probability to sample the neighbor of a point from the same manifold, and 
$1-\xi$  the probability  to sample it from a different manifold, with $\xi > 0.5.$
Define $n_{i}^{in} $ as the number of neighbors of $i$
with the same manifold membership ($n_{i}^{in} \equiv \sum_j \mathcal{N}_{ij}^{(q)} \delta_{z_i z_j}$). Then
we  introduce a probability distribution for 
$\boldsymbol{\mathcal{N}}^{(q)}$ as:
\begin{equation}
P(\boldsymbol{\mathcal{N}}^{(q)}|\pmb{z})=\prod_{i=1}^N \frac{\xi^{n_{i}^{in}}(1-\xi)^{q -n_{i}^{in}}}{\mathcal{Z}(\xi,N_{z_i})},
\end{equation}
where $\mathcal{Z}_q$ is a normalization
factor that depends also on the sizes of the manifolds (see SI for its explicit expression). This term favors homogeneity within a $q$-neighborhood of each point. 
With this addition, the model now reads:
\begin{equation}
\begin{split}
P_{post}(\pmb{z},\pmb{d},\pmb{p}| \pmb{\mu}, \mathcal{N}^{(q)}) 
\propto P( \boldsymbol{\pmb{\mu}} | \pmb{z},\pmb{d})  P( \boldsymbol{\mathcal{N}}^{(q)}|\pmb{z}) 
P_{pr}(\pmb{z}|\pmb{p})P_{pr}(\pmb{d})P_{pr}(\pmb{p}).\\
\end{split}
\label{eq:post2}
\end{equation}
The posterior (\ref{eq:post2}) is sampled with Gibbs sampling starting from a random configuration of the parameters. We repeat the sampling $M$ times starting from different random configurations of the parameters, keeping the chain with the highest maximum log-posterior value. The parameters $\pmb{d}$ and $\pmb{p}$ can be estimated by their posterior averages. For all estimations, we only include the last $10\%$ of the points. 
This ensures that the initial burn-in period is excluded \cite{diebolt1994estimation}. Moreover, visual inspection of the MCMC output is performed to ensure that we do not incur in any label switching issues \cite{Celeux1998}. 
Indeed, we observed that once the chains converge to one mode of the posterior distribution they do not transition to different modes. 
As for the $\pmb{z}$, we estimate the value of $\pi_{ik} \equiv P_{post}(z_i = k) $. Point $i$ can be safely assigned to manifold $k$ if
$\pi_{ik} > 0.8 $, otherwise we will consider its assignment to be uncertain. 
Note that our definition of ``manifold'' requires compactness: two spatially separated regions with approximately the same dimension are recognized as distinct manifolds by our approach.  
\\
We name our method Hidalgo (Heterogeneous Intrinsic Dimension Algorithm). Hidalgo has three free parameters: the number of manifolds, $K$; the local homogeneity range, denoted by $q$; the local homogeneity level, denoted by $\xi$.We fix $q$ and $\xi$ based on preliminary tests conducted on several artificial data sets, which show that the optimal ``working point'' of the method is at $q=3,\xi=0.8$ (SI, fig.~S1). As is common for mixture models, the value of $K$ can be estimated by model selection. 
In particular, we can compare the models with $K$ and $K+1$ for increasing $K$, starting from $K=1$ and stopping when there is no longer significant improvement, as measured  by the average log-posterior. \\
The computational complexity of the algorithm depends on the total number of points $N$, the number of manifolds $K$, and the parameter $q$. As input to the Gibbs sampling, the method requires the neighbor matrix and the $\pmb{\mu}$ vector. To compute the latter, one must identify the first $q$ neighbors of each point, which, starting from the distance matrix, requires
$O(q N^2)$ steps (possibly reducible to $O(q N \log N)$ with efficient nearest-neighbor-search algorithms~\cite{preparata2012computational}). The computation of a distance matrix from a coordinate matrix takes as $O(D N^2)$ steps, where $D$ is the total number of coordinates. In all practical examples with $N\lesssim 10^6$ points these preprocessing steps require much less computational time than the Gibbs sampling. The complexity of the Gibbs sampler is dominated by the sampling of the $\pmb{Z}$, which scales linearly with $N$ and $K$, and quadratically with $q$ (in total, $KNq^2$). The number of iterations required for convergence cannot be simply estimated. In practice, we found this time to be roughly linear in $N$ (SI, Fig~S2). Thus, in practice the total complexity of Hidalgo is roughly quadratic in $N$. The code implementing Hidalgo is freely available at https://github.com/micheleallegra/Hidalgo.

\section*{Results}

\subsubsection*{Validation of the method on artificial data}

\begin{figure*}
\includegraphics[width=\textwidth]{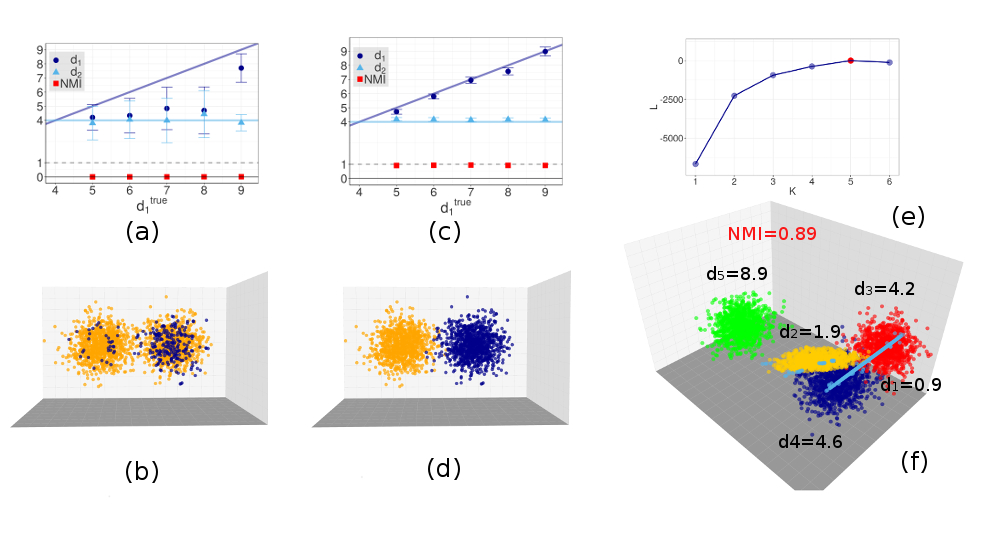} 
\vspace{-20pt}
\caption{
\textbf{Results on simple artificial data sets.} We consider sets of points drawn from mixtures of multivariate Gaussians in different dimensions.
In all cases, we perform $10^5$ iterations of the Gibbs sampling and repeat the sampling $M=10$ times starting from different random configurations of the parameters. We then consider the sampling with the highest maximum average of the log-posterior value.
\textbf{Panels a)-b)}:  Points are drawn from two Gaussians in different dimensions.
The higher dimension varies from $d_1=5$ to $d_1=9$, the
lower dimension is fixed at $d_2=4$}. $N=1000$ points are sampled from each manifold. We fix $K=2$. We show results obtained with $\xi=0.5$, namely, without enforcing neighborhood uniformity
(here $q=1$, but since $\xi=0.5$ the value of $q$ is irrelevant). In panel a) we plot the estimated dimensions of the manifolds (dots: posterior means; error bars: posterior standard deviations) and the MI between our classification and the ground truth.
In panel b) we show the assignment of points to the low-dimensional (blue) and high-dimensional (orange) manifolds for the case $d_1=9$ (points are projected onto the first 3 coordinates). Similar figures are obtained for other values of $d_1$.
\textbf{Panels c)-d)}: The same setting as in panels a)-b), but now we enforce neighborhood uniformity, using $\xi=0.8$ and $q=3$. Points are now correctly assigned to the manifolds whose ID is properly estimated.
\textbf{Panels e)-f)}: Points  drawn from five Gaussians in dimensions 
$d_1=1$, $d_2=2$, $d_3=4$, $d_4=5$, $d_5=9$.
$N=1000$ points are sampled from each manifold. 
Some pairs of manifolds are intersecting, as their centers are  one standard deviation apart. The analysis is performed assuming $\xi=0.8$, $q=3$, and with different values of $K$. In panel e) we show the average log-posterior value $\mathcal{L}$  
as a function of $K$. The maximum $\mathcal{L}$ corresponds to the ground truth value $K=5$. In panel f) we show the assignment of points to the five manifolds
in different colors (points were projected onto the first 3 coordinates).
\label{Fig: simulated}
\end{figure*}

\begin{figure*}
\includegraphics[width=\textwidth]{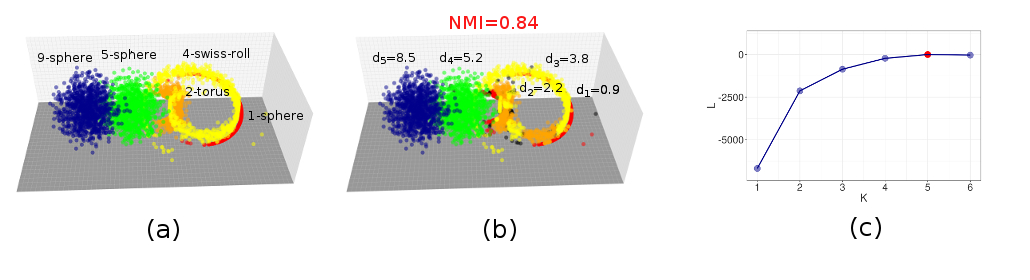}
\caption{\textbf{Results on data sets on curved manifolds.} We consider sets of points drawn from mixtures of multivariate Gaussians 
embedded in  curved nonlinear manifolds: a 1-dimensional circle, a 2-dimensional torus, a 4-dimensional Swiss roll, a 5-dimensional sphere, and a 9-dimensional sphere. All the manifolds are embedded in a 9-dimensional space. $N=1000$ points are sampled from each manifold.  Some pairs of manifolds are intersecting. In all cases, we perform $10^5$ iterations of the Gibbs sampling and repeat the sampling $M=10$ times starting from different random configurations of the parameters. We consider the sampling with the highest maximum log-posterior value. 
The analysis is performed with $\xi=0.8$, $q=3$, and with different values of $K$. In panel a) we show the ground-truth assignment of points to the five manifolds in different colors (points were projected onto the first 3 coordinates).  In panel b) we show the assignment of points as given by Hidalgo with $K=5$. In panel c) we show the average log-posterior value $\mathcal{L}$  as a function of $K$. The maximum $\mathcal{L}$ corresponds to the ground truth value $K=5$. }
\label{Fig: curved}
\end{figure*}

We first test Hidalgo on artificial data for which the true manifold partition of the data is known. 
We start from the simple case of two manifolds with different IDs, $d_1$ and $d_2$. We consider several examples, varying the higher dimension $d_1$ from $5$ to $9$ while fixing the lower dimension $d_2$ to $4$. On both manifolds $N=1000$ points are sampled from a multivariate Gaussian with a variance matrix given by $1/d_i$, $i=1,2$ multiplied by the identity matrix of proper dimension. The two manifolds are embedded in a space with dimension corresponding to the highest dimension $d_2$, with their centers at a distance of $0.5$ standard deviations, so they are partly overlapping. 
In Fig.~\ref{Fig: simulated}a-b we illustrate the results obtained in the case of fixed $\xi=0.5$, equivalent to the absence of any statistical constraint on neighborhood uniformity (note that for $\xi=0.5$ the parameter $q$ is irrelevant). 
The estimates of the two dimensions are shown together with the normalized mutual information (NMI) between the estimated assignment of points and the true one. The latter is defined as the MI of the assignment and the ground truth labeling divided by the entropy of the ground truth labeling.
As expected, without a constraint on the assignment of neighbors, the method is not able to correctly separate  the points and thus to estimate the dimensions of the two manifolds, even in the case of quite different IDs (NMI$<0.001$). As soon as we consider $\xi > 0.5$, results improve. A detailed analysis of the influence of the hyperparameters $q$ and $\xi$ is reported in SI. Based on such analysis, we identify the optimal parameter choice as $q=3,\xi=0.8$.
In Fig.~\ref{Fig: simulated}c-d we repeat the same tests as in \ref{Fig: simulated}a-b but with $q=3$ and $\xi=0.8$. Now the NMI between the estimated and ground truth assignment is almost $1$ in all cases and, correspondingly, the estimation of $d_1$ and $d_2$ is accurate.
To verify whether our approach can discriminate between more than two manifolds ($K > 2$), we consider a more challenging scenario consisting of five Gaussians with unitary variance in dimensions $1,2,4,5,9$ respectively. Some of the Gaussians have similar IDs, as in the case of dimensions $1$ and $2$, or $4$ and $5$: Moreover, they can be very close to each other: the centers of those in dimensions $4$ and $5$ are only half a variance far from each other, and they are crossed by the Gaussian in dimension $1$. To analyze such dataset we again choose the hyperparameters $q=3$ and $\xi=0.8$. We do not fix the number of manifolds $K$ to its ground truth value $K=5$, but we try to let the method estimate $K$ without relying on a priori information.
We perform the analysis with different values of $K=1,\dots,6$ and compute an estimate of the average of the logarithm of the posterior value $\mathcal{L}$ for each $K$.
Results are shown in Fig.~\ref{Fig: simulated}e. We see that $\mathcal{L}$ increases up to $K=5$, and then decreases, from which we infer that the optimal number of manifolds is $K=5$. In Fig.~\ref{Fig: simulated}f we illustrate the final assignment of points to the respective manifolds together with the estimated dimensions, upon setting the number of manifolds to $K=5$. The separation of the manifolds is very good. Only a few points of the manifold with dimension $1$ are incorrectly assigned to the one with dimension $2$ and vice versa. The values of normalized mutual information between the ground truth and our classification is 0.89.\\ 
Finally, we show that our method can handle data embedded in curved manifolds, or with complex topologies. Such data should not pose additional challenges to the method: Hidalgo is sensitive only to the local structure
of data. In terms of the distribution of the first and second nearest-neighbor distances,
a 2-d torus looks exactly like a 2-d plane, 
and thus we should be able to correctly classify these objects, that are so topologically different, as 2-d objects regardless of their shape. We generated a dataset with 5 Gaussians in dimension 1, 2, 4, 5, 9
(Fig.~\ref{Fig: curved}a). Each Gaussian is then embedded on a curved nonlinear manifold:
a 1-dimensional circle, a 2-dimensional torus, a 4-dimensional Swiss
roll, a 5-dimensional sphere, and a 9-dimensional sphere. None of these manifolds is topologically isomorphic to a hyperplane, except for the Swiss roll. Moreover, the manifolds are intersecting. As shown in  Fig.~\ref{Fig: curved}b, our approach can distinguish the five manifolds. The dimensions are correctly retrieved (Fig.~\ref{Fig: curved}b), points are assigned to the right manifold with an accuracy corresponding to a value of the normalized mutual information (NMI) $=$0.84. Only some points of the 3d-torus are misassigned to the 4d-swiss roll. As in the previous example, $\mathcal{L}$ increases up to $K=5$, and then decreases, from which we infer that the optimal number of manifolds is $K=5$ ( Fig.~\ref{Fig: curved}c).

\subsubsection*{ID variability in a protein folding trajectory}

\begin{figure}
\centering
\includegraphics[width=0.5\columnwidth]{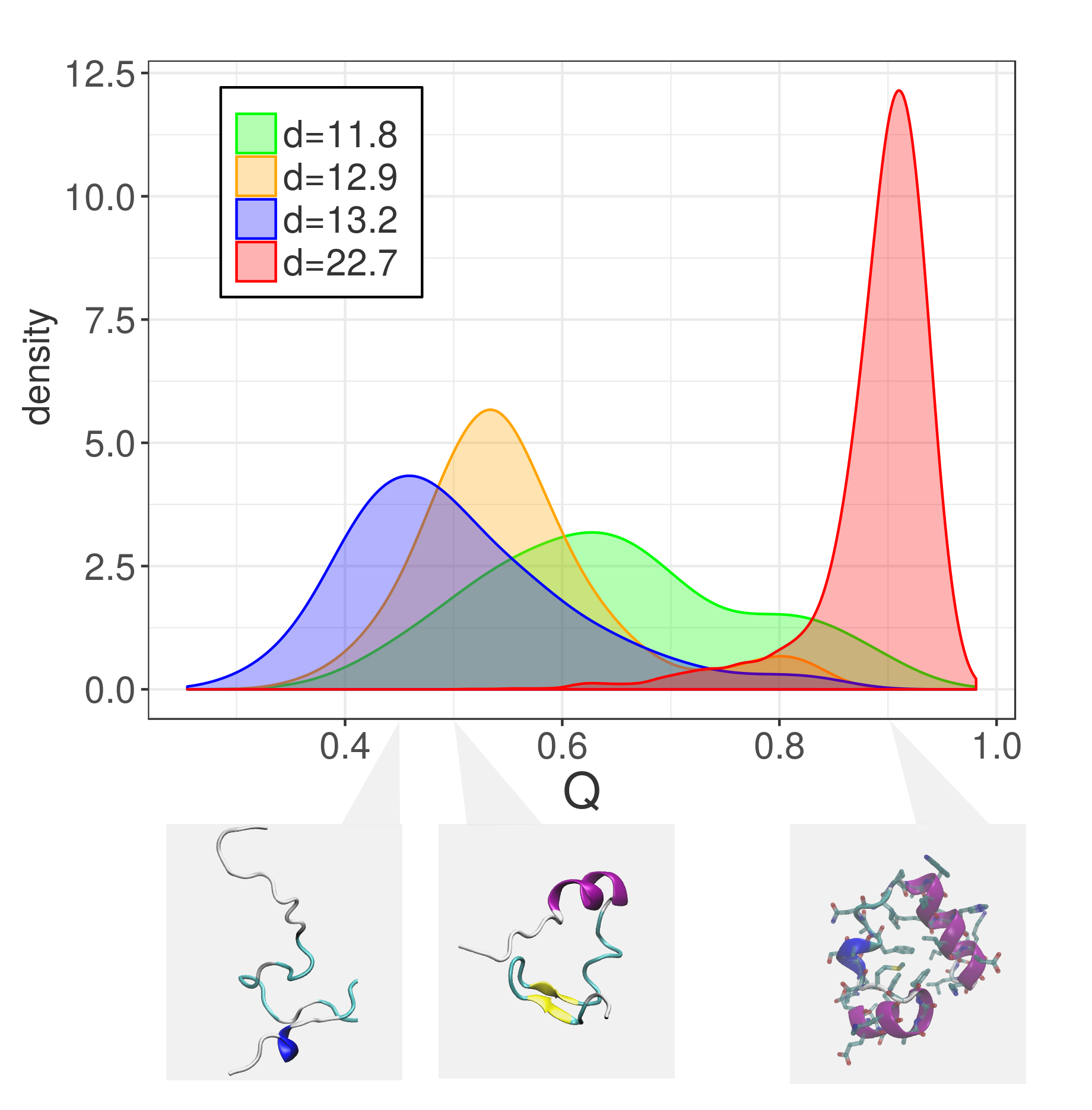}
\caption{\textbf{Protein folding trajectory}. We consider $N \sim 31,000$ configurations of a protein undergoing successive folding/unfolding cycles. For each configuration, we extract the value of the $D=32$ 
 backbone dihedral angles.  Applying Hidalgo to these data, we detect four manifolds, of intrinsic dimensions $11.8$, $12.9$, $13.2$, and $22.7$. For each configuration, we also compute the fraction of native contacts, $Q$, which  
measures to which degree the configuration is folded.
The figure shows the probability distribution of $Q$ in each manifold.  
Most of the folded configurations belong to the high-dimensional manifold: the  analysis essentially identifies the folded configurations as a region of high intrinsic dimension. 
Results are obtained  with $q=3$ and  $\xi=0.8$. The distance between each pair of configurations was computed by the Euclidean metric with periodic boundary conditions on the vectors of the dihedral angles.}
\label{Fig: folding}
\end{figure}

\begin{figure}
\includegraphics[width=\columnwidth]{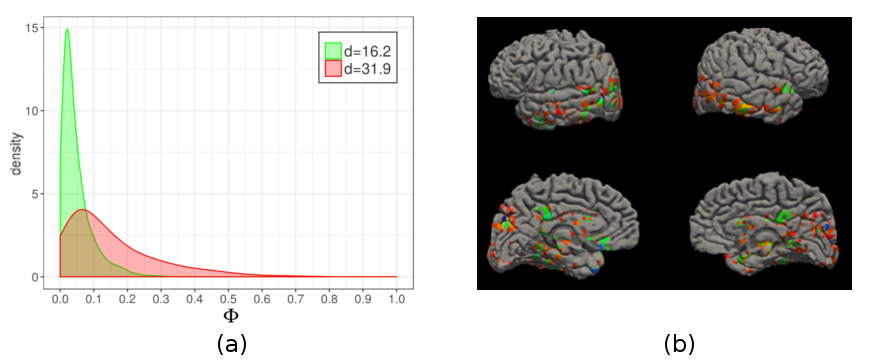}	
\caption{\textbf{Neuroimaging data.} 
We consider the BOLD time series of $N \sim 30,000$ voxels in an fMRI experiment with $D=202$ scans. Hidalgo detects a low-dimensional manifold ($d = 16.2$) and a high-dimensional one ($d= 31.9$). We compute the clustering frequency $\Phi$, which measures the participation of each voxel to coherent activation patterns and is a proxy for voxel involvement in the task \cite{allegra2020brain}. 
Panel (a) shows the probability distribution of  $\Phi$ in the two manifolds. Strongly activated voxels ($\Phi > 0.2$) are consistently assigned to the high-dimensional manifold. Panel (b) shows the rendering of the cortical surface (left: left hemisphere; right: right hemisphere). Blue voxels have high clustering frequency ($\Phi > 0.2$),  red voxels are those assigned to the high-dimensional manifold, and green voxels satisfying both criteria. 
Almost all voxels with high clustering frequency are assigned to the high-dimensional manifold, and are concentrated in the occipital, temporal, and parietal cortex.
Results are obtained with $q=3$ and $\xi=0.8$. The distance between two time series is computed by a Euclidean metric, after standard pre-processing steps\cite{allegra2017fmri}.}
\label{Fig: fmri}
\end{figure}

\begin{figure}
\includegraphics[width=\columnwidth]{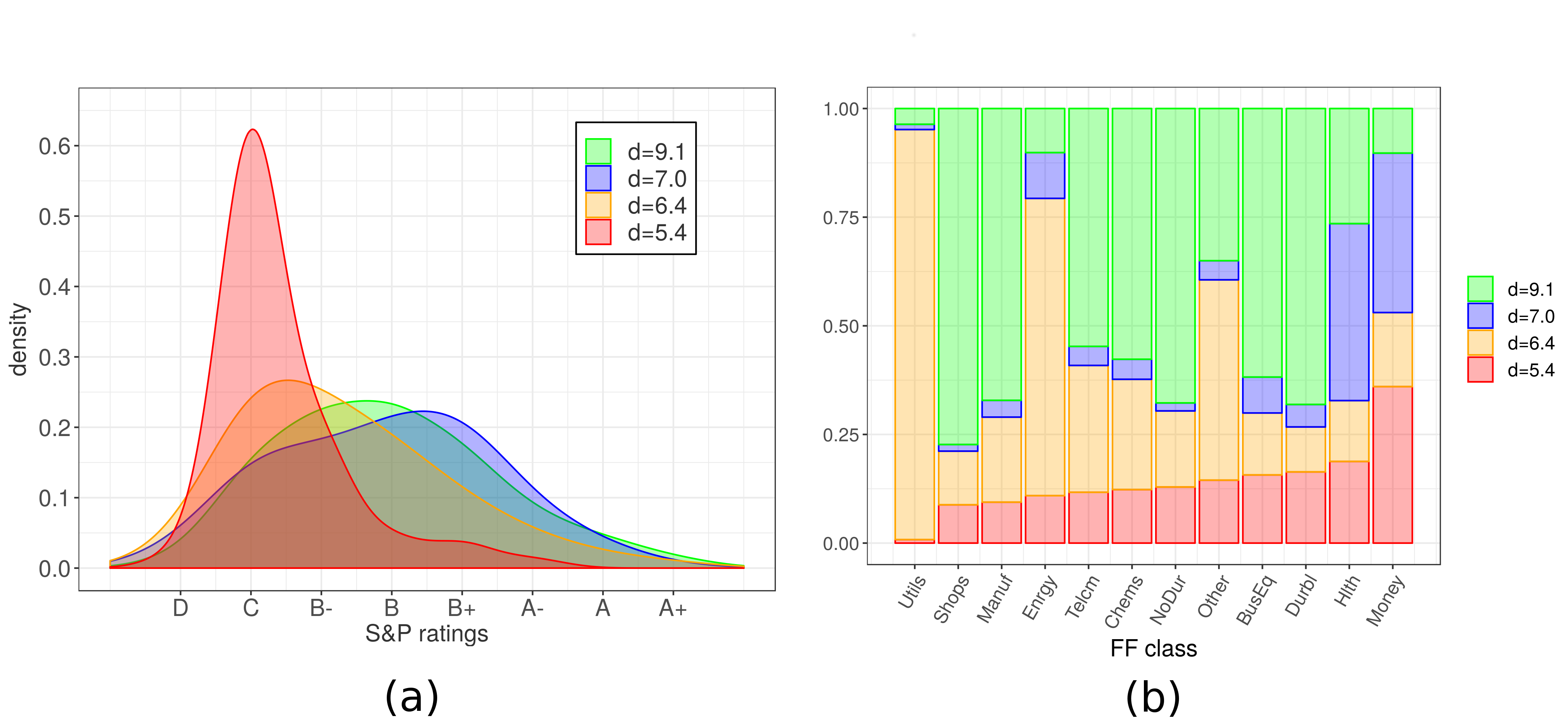}
\caption{\textbf{Financial data.}
For $N \sim 8,000$ firms selected from the COMPUSTAT database, we compute $D=31$ variables from their yearly balance sheets.  Hidalgo finds four manifolds of intrinsic dimensions $5.4$, $6.4$, $7.0$, and $9.1$. Panel (a) shows the fractions of firms assigned to the four manifolds for each type of firm, according to the Fama-French classification. The four manifolds contain unequal proportions of firms belonging to different classes, implying that some classes of firms are preferentially assigned to manifolds of high vs low dimension. Panel (b) shows the probability distribution of the S\&P ratings of the firms assigned to each manifold. Firms with low ratings preferentially belong to low-dimensional manifolds. Results are obtained  with $q=3$ and $\xi=0.8$. To correct for firm size, we normalize the variable vector of each firm by its norm, and then applied standard Euclidean metric.
}
\label{Fig: financial}
\end{figure}
As a first real application of Hidalgo, we address ID estimation for a dynamical system. Asymptotically, dynamical systems are usually restrained to a low-dimensional manifold in phase space, called an attractor. Much effort has been devoted to characterizing the ID of such attractor~\cite{grassberger2004measuring}. 
However, in the presence of multiple 
metastable states 
an appropriate description of the visited phase space may require the use of multiple IDs.
Here, we consider the dynamics of the villin headpiece (PDB entry: 2F4K). 
Due to its small size and fast folding kinetics,
this small protein is a prototypical system for molecular dynamics simulations. Our analysis is based on the longest available simulated trajectory of the system from Ref. \cite{lindorff2011fast}. 
During the simulated $125$ $\mu s$, the protein performs approximately 10 transitions between the folded and the unfolded state. 
We expect to find different dimensions in the folded and unfolded state, since these two states are metastable, and they would be considered as different attractors in the language of dynamical systems. Moreover, they are characterized by different chemical and physical features: the folded state is compact and dry in its core, while the unfolded state is swollen, with most of the residues interacting with a large number of water molecules. 
We extract the value of the 32 key observables (the backbone dihedral angles) for all the $N=31,000$ states in the trajectory and apply Hidalgo to this data of extrinsic dimension $D=32$. We obtain a vector of estimated intrinsic dimensions $\pmb{d}$ and
an assignment 
of each point $i$ to one of the $K$ manifolds. 
We find four manifolds, three low-dimensional ones ($d_1 = 11.8$, $d_2 = 12.9$, $d_3 = 13.2$) 
and a high-dimensional one ($d_4 = 22.7$). We recall that two spatially separated regions with approximately the same dimension (in this case, $d_2$ and $d_3$) are recognized as distinct manifolds by our approach.  
To test whether this partition into manifolds is related to the separation between the folded and the unfolded state we relate the partition to the fraction of native contacts $Q$, which can be straightforwardly estimated on each configuration of the system. $Q$ is close to one only if the configuration is folded, while it approaches zero when the protein is unfolded. In Fig.~\ref{Fig: folding} we plot the probability distribution of $Q$ restricted to the four manifolds. 
We find that the vast majority of  the folded configurations ($Q>0.8$) are assigned to the high-dimensional manifold. Conversely, the unfolded configurations ($Q\leq0.8$) are most of the times assigned to one of the low-dimensional manifolds. This implies that a configuration belonging to the low dimensional manifolds is almost surely unfolded.
The normalized mutual information  between the manifold assignment and the folded/unfolded classification is 0.60. Thus, we can essentially identify the folded state using the intrinsic dimension, a purely topological observable unaware of any chemical detail. 

\subsubsection*{ID variability in time-series from brain imaging}

In the next example, we analyze a set of time-series from functional resonance imaging (fMRI) of the brain, representing the BOLD (blood oxygen-level dependent) signal of each voxel, which captures the activity a small part of the brain~\cite{huettel2004functional}. 
fMRI time-series 
are often  projected on a lower dimension through linear projection techniques like PCA~\cite{poldrack2011handbook}, a step that assumes a uniform ID. However, the gross features of the signal (e.g., power spectrum and entropy) are often highly variable in different parts of the brain, and
also non-uniformities in the ID may well be present.
Here, we consider a single-subject fMRI recording containing $D=202$ images collected while a subject was performing a visuomotor task~\cite{schuck2015medial,allegra2020brain}. From the images we extract the $N=29851$ time series corresponding to the BOLD signals of each voxel.
Applying Hidalgo, we find two manifolds 
with very different dimensions $ d_1 = 16.2$, $d_2 = 31.9$. 
Again, we relate the identified manifolds to a completely independent quantity, the clustering frequency $\Phi$ introduced in \cite{allegra2017fmri,allegra2020brain}, which measures the temporal coherence of the signal of a voxel  with the  signals of other voxels in the brain.
Voxels with non-negligible clustering frequency ($\Phi > 0.2 $) are likely to belong to  brain areas involved in the cognitive task at hand.  In Fig.~\ref{Fig: fmri} we show the probability distribution of $\Phi$ restricted to the two manifolds.  
We find that the ``task-related''  voxels ($\Phi > 0.2$) almost invariably belong to the manifold with high dimensionality.
These voxels appear concentrated in the occipital, parietal, and temporal cortex (Fig.~\ref{Fig: fmri} bottom), and belong to a task-relevant network of coherent activity~\cite{allegra2020brain}. 
The normalized mutual information  between the manifold assignment and the task-relevant/task-irrelevant classification is 0.24; note that this value is lower than for the protein folding case, because the the low-dimensional manifold is ``contaminated'' by many points with low $\Phi$. 
This result finds an interesting interpretation:  the subset of ``relevant'' voxels gives rise to patterns that are not only coherent but also characterized by a larger ID than the remainder of the voxels. On the contrary, the incoherent voxels exhibit a lower ID, hence a reduced variability, which is consistent with the fact that the corresponding time series are dominated by low-dimensional noise. Again, this feature emerges from the global topology of the data, revealed by our ID analysis, without exploiting any knowledge of the task that the subject is performing.

\subsubsection*{ID variability in financial data}

Our final example is in the realm of economics. 
We consider firms in the well-known Compustat database ($N=8309$). For each firm, we consider $D=31$ balance sheet variables from the fiscal year 2016 (for details, see Table S2 in the SI). Applying Hidalgo we find 
four manifolds of dimensions $d_1=5.4$, $d_2=6.3$, $d_3=7.0$ and $d_4=9.1$. 
To understand this result, we try to relate our classification with common indexes showing the type and financial stability of a firm.
We start by relating our classification to the Fama-French classification~\cite{fama1997industry}, which assigns each firm to one of twelve categories depending on the firm's trade. In Fig.~\ref{Fig: financial}(top) we separately consider firms belonging to the different Fama-French classes, and 
compute the fraction of firms assigned to each the four manifolds 
identified by Hidalgo.
The two classifications are not independent, 
since the fractions for different Fama-French classes are highly non-uniform.
More precisely, the normalized mutual information between the two classifications is $0.19$, rejecting hypothesis of statistical independence ($p$-value $< 10^{-5}$). 
In particular, firms in the utilities and energy sector show a preference for low dimensions ($d_1$ and $d_2$), 
while firms purchasing products (nondurables, durables, manufacturing chemicals, equipment, wholesale) are concentrated in the manifold with the highest dimension $d_4$. The manifold with intrinsic dimension $d_3$ mostly includes firms in the financial and health care sectors. 
Different dimensions are not only related to the classification of the firm, but also their financial robustness. 
We consider the S\&P quality ratings for the firms assigned to each manifold (also from Compustat; ratings are available for $2894$ firms). In Fig.~\ref{Fig: financial}(bottom) we show the distribution of ratings for the different manifolds. These distributions appear to be different. We cannot predict the rating based on the manifold assignment (the normalized mutual information between the manifold assignment and the rating is 0.04), but companies with worse ratings show a preference to belong to low-dimensional manifolds: by converting ratings into numerical values scores from $1$ to $8$, we found a Pearson correlation of $0.22$ between the local ID and the rating ($p<10^{-10}$) 
We suggest a possible interpretation for this phenomenon: a low ID may imply a more rigid balance structure, which may entail a higher sensitivity to market shocks which, in turn, may trigger domino effects in contagion processes. 
This result shows that information about the S\&P rating can be derived using only the topological properties of the data landscape, without any in-depth financial analysis.
For example, no information on the commercial relationship between the firms or the nature of their business is used.

\subsubsection*{Comparison with other methods}

\begin{table}[h!]
    \centering
    \small
        \begin{tabular}{| C{1.8cm} |  C{1.2cm} |  C{2.5cm} | C{0.7cm} | C{3.3cm} | C{0.7cm} |  C{1.8cm} | C{0.7cm} | }
    \hline
        \textbf{Dataset} & & \textbf{Hidalgo} \qquad \qquad  d &  \qquad \qquad \qquad NMI & \textbf{Ref. \cite{elhamifar2011sparse}} \qquad \qquad \qquad  d   & \qquad \qquad \qquad NMI & \textbf{Ref. \cite{carter2010local}} \qquad \qquad \qquad  d   & \qquad \qquad \qquad NMI  \\
            \hline
      2 Gaussians & $d_1=5$ $d_2=4$ & $d_1=4.7$ \qquad $d_2=4.2$  & 0.91 & $d_1=5$ \qquad \quad \qquad  $d_2=4$ & 0.96 &  $\langle d \rangle = 4.0$ \qquad    $\langle d \rangle = 4.7$    & 0.44 \\ 
       \hline
          2 Gaussians & $d_1=6$ $d_2=4$ & $d_1=5.8$ \qquad $d_2=4.2$  & 0.93 & $d_1=6$  \quad \qquad \qquad $d_2=4$ & 0.94  & $\langle d \rangle = 4.0$ \qquad    $\langle d \rangle = 5.1$ &   0.95 \\
              \hline
          2 Gaussians & $d_1=7$ $d_2=4$ & $d_1=6.9$ \qquad $d_2=4.1$  & 0.94  & $d_1=6$  \quad \qquad \qquad $d_2=4$ & 0.97  & $\langle d \rangle = 4.0$ \qquad    $\langle d \rangle = 6.0$  &   0.96 \\
              \hline
          2 Gaussians & $d_1=8$ $d_2=4 $ &  $d_1=7.6$ \qquad $d_2=4.2$ & 0.92 & $d_1=7$   \quad \qquad \qquad $d_2=4$ & 0.94  & $\langle d \rangle = 4.0$ \qquad    $\langle d \rangle = 6.7$ &    0.93 \\
           \hline
           2 Gaussians & $d_1=9$ $d_2=4 $ &  $d_1=9.0$ \qquad $d_2=4.1$ & 0.92 & $d_1=7$   \quad \qquad \qquad $d_2=4$ & 0.95  & $\langle d \rangle = 4.0$ \qquad    $\langle d \rangle = 7.2$ &   0.92 \\
           \hline
           5 Gaussians (linear) &  $d_1=1$ $d_2=2$ $d_3=4$ $d_4=5$ $d_5=9$  & $d_1=0.9$ \qquad $d_2=1.9$ \qquad $d_3=4.2$ \qquad $d_4=4.6$ \qquad $d_5=8.9$  & 0.89 & $d_1=7,d_4=9,d_5=10$   $d_2=10$  \qquad \qquad \qquad   $d_3=9$ \qquad \quad \qquad  $d_4=9$ \qquad \qquad \quad \ \  $d_5=10$ & 0.73  & $\langle d \rangle = 1.0$ \qquad    $\langle d \rangle = 2.0$  \quad $\langle d \rangle = 3.5$ \qquad    $\langle d \rangle = 4.3$  \qquad    $\langle d \rangle = 7.3$    & 0.81 \\
          \hline
          5 Gaussians (curved) & $d_1=1$ $d_2=2$ $d_3=4$ $d_4=5$ $d_5=9$  & $d_1=0.9$   $\qquad \qquad d_2=2.2   $ $d_2=2.2,d_3=3.8$ $d_4=5.2$ \qquad $d_5=8.5$  & 0.84 & $d_1=6,d_3=7$ \qquad $d_2=10$ \qquad \qquad \quad   $d_3=7,d_4=8$  \qquad \qquad  $d_5=10$ \qquad \qquad  \quad  $d_5=10$ & 0.61  & $\langle d \rangle = 1.0$ \qquad    $\langle d \rangle = 2.0$  \quad $\langle d \rangle = 2.0$ \qquad    $\langle d \rangle = 4.6$  \qquad    $\langle d \rangle = 7.2$ & 0.77 \\
          \hline
          Protein Folding & $Q \leq 0.8$ \qquad \qquad $_{}$ \qquad \qquad \qquad $_{}$ \qquad \qquad \qquad  $Q > 0.8$  &  $d_1=11.8$ $d_2=12.9$ $d_3=13.2$ $d_4=22.3$   \qquad \qquad & 0.60  & $d_1=11$  & 0  &  $\langle d \rangle = 11.4$  \qquad   
          ${}_{}$  \qquad \qquad \qquad  ${}_{}$  \qquad
          $\langle d \rangle = 13.7$  & 0.22  \\
          \hline
    \end{tabular}
    \caption{\textbf{Performances of Hidalgo, restricting analysis to points with non-overlapping first and second neighbor}  For each data set we show the number of points in the original data set ($N_0$), the number of points upon restriction $N$, the estimated value of the $d_k$, and NMI betzeen the assignment and the ground truth. Qll results were obtained with $\xi=0.8,q=3$, repeating the sampling $M=10$ times and considering the sampling with the highest average log-posterior. }
    \label{tab:comp0}
\end{table}

\begin{figure*}[h!]
\includegraphics[width=0.9\textwidth]{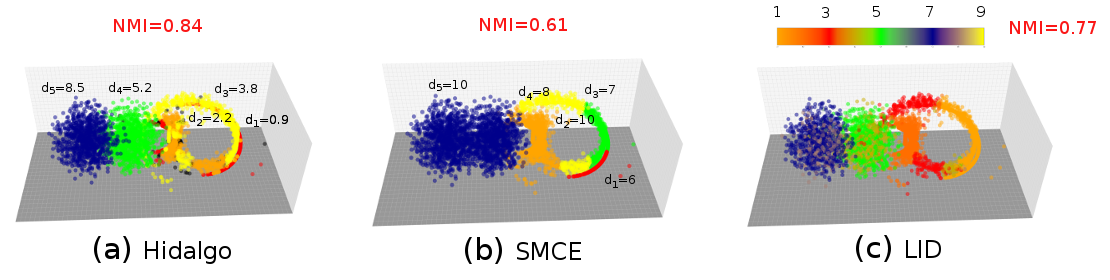} 
\caption{\textbf{Results of other variable-ID analysis methods on topologically complex artificial data}. We consider sets of points drawn from mixtures of multivariate Gaussians embedded in  curved nonlinear manifolds, as detailed in fig.~\ref{Fig: curved},
and compared the results of Hidalgo(a), SMCE (b), LID (c). In panel a) we plot again, for ease of comparison, the results of Fig.~\ref{Fig: curved}b. In panel b), we show the local ID estimates given by LID~\cite{carter2010local}, considering $k=50$ nearest neighbors for local ID estimation and smoothing. The method assigns an integer dimension to each point: in the plot, the dimension of each point is represented by its color, according to the color bar shown. The method correctly retrieves the manifolds of dimension 5 and 9, even though the points in the manifold with dimension 5 are assigned a local ID that oscillates between 4 and 5, and points in the manifold with dimension 9 are assigned a local ID that oscillates between 7 and 8. Also, the manifold of dimension 2 is well identified. The manifolds of dimension 1 and 4,  however, cannot be correctly discriminated. Points of the manifold of dimension 4 are assigned ID estimates of 1 or 3, without a clear separation of the manifolds. The NMI between the ground truth and the integer local ID is 0.77. In panel c) we show the assignment of points as given by SMCE~\cite{elhamifar2011sparse}. The manifolds of dimension 5 and 9 are merged. The manifold of dimension 2 is identified, but it is also contaminated by points from the manifold of dimension 4. SMCE cannot correctly retrieve the manifolds of dimension 1 and 4.  The NMI between the ground truth and the assignment is 0.61. ID estimates obtained according to the prescription given in~\cite{elhamifar2011sparse} are largely incorrect.}
\label{Fig: comparison}
\end{figure*}

We have compared our method with two state-of-the-art methods in the literature, considering the synthetic datasets as well as the protein folding dataset (for which the folded/unfolded classification yields an approximate ``ground truth''). The first method we consider is the ``sparse manifold clustering and embedding'' (SMCE, Ref.~\cite{elhamifar2011sparse}). The method creates a graph by connecting neighboring points, supposedly lying on the same manifold, and then uses spectral clustering on this graph to retrieve the manifolds. As an output, it returns a manifold assignment for each point, together with an ID estimate. We resort to the implementation provided by Ehsan Elhamifar (\url{http://khoury.neu.edu/home/eelhami/codes.htm}). For SMCE, we can compute the NMI between the manifold assignment and the ground-truth manifold label. Furthermore, we can report the estimated local ID for points with a given ground-truth assignment, identified with the ID of the manifold to which they are assigned. SMCE requires to specify the number of manifolds $K$ as input and depends on a free parameter $\lambda$.  To present a fair comparison, we fixed $K$ at its ground truth value and explored different values in the range $0.01 \leq \lambda \leq 20$, and we report the results corresponding to the highest NMI with the ground truth. Furthermore, we repeated the estimation $M=10$ times and kept the results with the highest NMI with the ground truth. The second method is the local ID (LID) estimation by Carter et al.~\cite{carter2010local}, which combines ID estimation restricted to a (large) neighborhood and local smoothing to produce an integer ID estimate for each point. We used the implementation provided by Kerstin Johnsson~\cite{johnsson2016structures} 
(\url{https://github.com/kjohnsson/intrinsicDimension}) using a neighborhood size  $k=50$ which is a standard value. Note that the LID estimation is deterministic, so we do not need to repeat the estimation several times. For LID, we can compute the NMI between the (integer) label given by the local ID estimate and the ground truth; moreover, we can compute the average ID, $\langle d \rangle $ for the points with a given ground truth assignment. \\
The results of the comparison are presented in Table \ref{tab:comp0}. For the examples with two manifolds, SMCE~\cite{elhamifar2011sparse} is able to correctly retrieve the manifolds with high accuracy (NMI$\geq 0.94$). The estimate of the lower dimension $d_2=4$ is always correct, while the highest dimension is underestimated for $d_1 \geq 7$.
Overall, SMCE performs very well (even better than Hidalgo) on the datasets with two manifolds, as the two manifolds are well separated, matching well the assumptions of the model. On the examples with five manifolds, the performance of SMCE is worse, as the method cannot correctly retrieve intersecting manifolds. In the example with five linear manifolds, SMCE cannot retrieve the manifold of dimension $d_1=1$, which is intersecting the manifolds of dimension $d_3=4$ and $d_4=5$. Moreover, the ID estimates are quite poor.  In the example with five curved manifolds, SMCE merges the manifolds with $d_4=5$ and $d_5=9$,
and cannot retrieve the manifolds with  $d_1=1$ and $d_3=4$, which get split. Again, the dimension estimates are poor. For the protein folding data, where regions with different IDs are probably not well separated, SMCE finds a single manifold with $d=11$, and thus it is not able to detect any ID variation. \\
For the examples with two manifolds, the LID method~\cite{carter2010local} is generally able to correctly retrieve regions with different IDs with high accuracy (NMI$\geq 0.92$), except for the case with $d_1=5,d_2=4$ where results are inaccurate. The estimate of the lower dimension $d_2=4$ is always precise, while the higher dimension is always underestimated.  The degree of underestimation increases with increasing $d_1$. This is expected, since the ID estimates are produced by assuming a uniform density in a large neighborhood of each point, an assumption that fails in these data, especially for points on the manifold with higher dimension. Overall, LID performs quite well (comparably with Hidalgo) in separating regions with different IDs in the datasets with two manifolds, but gives worse ID estimates than Hidalgo. In the examples with five manifolds, the performance of LID is slightly worse. In the example with five linear manifolds LID merges the manifolds of dimension $d_3=4$ and $d_=5$, and misassigns some of their points to the manifold of dimension $1$. This is because relatively large neighborhoods are used for the estimation of the local ID: this leads to a difficulty in discriminating close regions with similar ID ( $d_3=4$ and $d_4=5$) and to a ``contamination'' of results by points of the manifold with $d_1=1$ intersecting the two.  In the example with five curved manifolds, LID can correctly identify the manifolds with $d_2=2$, $d_4=5$, and $d_5=9$, even if it underestimates the ID in the latter two cases (high dimension, large density variations). Many points in the manifold of dimension $d_3=4$ are mis-assigned to the manifold of dimension $d_1=1$, due to the fact that the two manifolds are highly intersecting. Finally, for the protein folding data, where the density is highly variable, LID shows a tendency for unfolded configurations to have a higher local ID than unfolded ones. However, it cannot discriminate between folded and unfolded configurations as Hidalgo. The ID of the folded configurations is highly underestimated compared to the Hidalgo estimate, probably because of density variations. \\
In Figure \ref{Fig: comparison} we compare the results of the three methods (Hidalgo, SMCE, and LID) for the dataset with five curved manifolds.

\section*{Discussion}

The increasing availability of a large amount of data has considerably expanded the opportunities and challenges for unsupervised  data analysis. 
Often data come in the form of a completely uncharted
``point cloud'' for which no model is at hand.
The primary  goal of the analyst is to uncover some structure within the data. For this purpose, a typical approach is dimensionality reduction, whereby the data are simplified by projecting them onto a low-dimensional space. \\
The appropriate \emph{intrinsic dimension (ID)} of the space onto which one should project the data is not constant everywhere. In this work, we developed an algorithm (Hidalgo) to find manifolds of different IDs in the data. 
Applying Hidalgo, we observed large variations of the ID in datasets of diverse origin (a molecular dynamics simulation, a set of time series from brain imaging, a dataset of firm balance sheets).  This finding suggests that a highly non-uniform ID is not an oddity, but a rather common feature. Generally speaking, ID variations can be expected whenever the system under study can be in various ``macrostates'' characterized by a different number of degrees of freedom due, for instance, 
to differences in the constraints. 
As an example, the folded state the protein is able to locally explore the phase space in many independent directions, while in the unfolded
state it only performs wider movements in fewer  directions.
In the case of companies, a financially stable company may have more
degrees of freedom to adjust its balance sheet. \\ 
In the cases we have analyzed, regions characterized by different dimensions were found to host data points differing in important properties. Thus, variations of the ID within a dataset can be used to classify the data into different categories. 
It is remarkable that such classification is achieved by looking at a simple topological property, the local ID.
Let us stress that ID-based segmentation should not be considered an alternative to clustering methods. In fact, in most cases (e.g., most classical benchmark sets for clustering) different clusters do not correspond to significantly
different intrinsic dimensions - rather, they correspond to regions
of high density of points within a single manifold of well-defined
ID. In such cases, clusters cannot be identified as different manifolds by Hidalgo. Conversely, when the ID is variable, regions of different IDs do not necessarily correspond to clusters in the standard meaning of the word: they may or may not correspond to regions of high density of points. 
A typical example
could be that of protein folding: while the folded configurations
are quite similar to one another, and hence constitute a cluster in
the traditional sense, the unfolded configurations may be very heterogeneous,
hence quite far in data space and then not a ``cluster'' in the
standard meaning of the word.\\
The idea that ID may vary in the same data is not new. In fact, many works have discussed the possibility of a variable ID and developed methods to estimate multiple IDs~\cite{barbara2000using,gionis2005dimension,costa2005estimating,carter2010local,campadelli2013local,johnsson2015low,mordohai2005unsupervised,amsaleg2015estimating,faranda2017dynamical,haro2008translated,souvenir2005manifold,wang2010multi,goh2007segmenting,vidal2011subspace,elhamifar2013sparse}. 
Our method builds on these previous contributions but is designed with the specific goal of overcoming technical limitations of other available approaches and make  ID-based segmentation a general-purpose tool. 
Our scheme uses only the distances between the data points, and not their coordinates, which significantly enlarges its scope of applicability. Moreover, the scheme uses only the distances between a point and its  $q$ nearest neighbors, with $q \le 5$. We thus circumvent the notoriously difficult problem of defining a globally meaningful metric~\cite{tenenbaum2000global}, only needing a consistent metric on a small local scale.
Finally, Hidalgo is not hindered by density variations or curvature. 
For these reasons, Hidalgo is competitive with other manifold learning and variable ID estimation methods and, in particular, can yield better ID estimates and manifold reconstruction. We have compared our method with two state-of-the-art methods in the literature, the local ID method (LID, \cite{carter2010local}), and the sparse manifold clustering and embedding (SMCE, \cite{elhamifar2011sparse}) method. Both methods show issues in the cases of intersecting manifolds and variable densities and yield worse ID estimates than Hidalgo, especially for large IDs. \\
Hidalgo is computationally efficient and therefore suitable for the analysis of large data sets. For example, it takes $\approx$30 mins to perform the analysis of the neuroimaging data, which includes 30000 data points, on a standard computer using a single core. Implementing the algorithm with parallel Gibbs sampling~\cite{gonzalez2011parallel} may considerably reduce computing time and then yield a fully scalable 
method. 
Obviously, Hidalgo  has some limitations. 
Some are intrinsic to the way the data are modeled. Hidalgo is not suitable to cover cases in which the ID is a continuously varying parameter~\cite{amsaleg2015estimating}, or in which sparsity is so strong that points cannot be assumed to be sampled from a continuous distribution. In particular, Hidalgo is not suitable for discrete data for which the basic assumptions of the method are violated. For example, ties in the sample could lead to null distances between neighbors, jeopardizing the computation of $\pmb{\mu}$.
Others are technical issues related with the estimation procedure, and, and least in principle, susceptible to improvement in refined versions of the algorithm: for instance,
one may improve the estimation with suitable enhanced sampling techniques.
Finally, let us point out some further implications of our work. Our findings suggest a caveat with respect to common practices of dimensionality reduction, which assume a uniform ID. In the case of significant variations, a global dimensionality reduction scheme may become inaccurate. 
In principle, the  partition in manifolds obtained with Hidalgo may be the starting point for using standard dimensionality reduction schemes. For example, one can imagine to apply PCA\cite{jolliffe1986principal} or Isomap\cite{tenenbaum2000global}, or sketchmap\cite{ceriotti2011simplifying} separately to each manifold. However, we point out that a feasible scheme to achieve this goal does not come as an immediate byproduct of our method. Once a manifold with given ID is identified, it is highly nontrivial to provide a suitable parameterization thereof, especially because the manifolds may be highly nonlinear, and even topologically non-trivial. How to suitably integrate our approach with a dimensionality reduction scheme remains a topic for further research. Another  implication is that a simple topological invariant, the ID, can be very a powerful tool for unsupervised data analysis, lending support to current efforts at characterizing topological properties of the data~\cite{carlsson2009topology,zomorodian2005computing}.

\bibliography{bib}

\subsubsection*{Data availability} Figure 1, 2, 6 and 7 have raw simulated data. Figure 3,4,5 use real data.
All data used are publicly available in the following github repository: https://github.com/micheleallegra/Hidalgo

\subsubsection*{Code availability}
Code is publicly available in the following github repository: https://github.com/micheleallegra/Hidalgo

\subsubsection*{Acknowledgements}
 MA was supported by Horizon 2020 FLAG-ERA (European Union), grant ANR-17-HBPR-0001-02 during completion of this work.  
 We thank Giovanni Barone Adesi and Julia Reynolds (Institute of Finance, USI, Lugano, Switzerland) for helping us with the Compustat dataset and offering precious suggestions for its analysis.  We thank Giulia Sormani (SISSA, Trieste, Italy) for suggesting, and providing us with the protein dynamics data. We thank Alex Rodriguez (SISSA, Trieste, Italy) for helping with the analysis of protein data. 
M.A. thanks the SISSA community at large for the substantial moral and intellectual support received, which has been critical for completion of this work.\ \\
 The authors declare that they have no
competing financial interests. \ \\
 Correspondence and requests for materials should be addressed to A.L.~(email: laio@sissa.it) and A.M..~(email: antonietta.mira@usi.ch).

\newpage

\renewcommand{\thetable}{S\arabic{table}}  
\renewcommand{\thefigure}{S\arabic{figure}}
\renewcommand{\theequation}{S\arabic{equation}}
\setcounter{figure}{0} 
\setcounter{table}{0} 
\setcounter{equation}{0} 

\section*{Supplementary Information}

\textbf{Enforcing neighborhood uniformity}. In our model, we wish to obtain well separated manifolds. We do not wish to impose this condition in the form of a rigid constraint, since in real cases regions with different IDs are not completely separated, but only as a ``soft constraint'', privileging configurations of $\pmb{z}$ such that the first neighbors of each point are preferentially assigned to the same manifold. In a Bayesian framework, this means that given that $j$ is among the first neighbors of $i$, the probability that $z_i = z_j$ is increased.
Consider the \emph{neighbor matrix} $\mathcal{N}_{ij}^{(q)}$ defined as: 
\begin{equation}
\mathcal{N}_{ij}^{(q)}=\begin{cases}
1 & \mbox{if }j\neq i\ \mbox{\mbox{is among the first \ensuremath{q} neighbors of }}i\\
0 & \mbox{otherwise, including } i=j
\end{cases}
\end{equation}
 \\ 
Intuitively, we would like to impose
\begin{equation}
P_{post}(z_i = z_j | \mathcal{N}_{ij}^{(q)}=1, \boldsymbol{\mu},\boldsymbol{p}) > P_{post}(z_i = z_j | \mathcal{N}_{ij}^{(q)}=0,\boldsymbol{\mu},\boldsymbol{p}).
\label{Eq: uniformitypost}
\end{equation}
\\ 
However, Eq. \ref{Eq: uniformitypost} is a relation between posterior probabilities, hence it cannot be directly embedded in the likelihood. What we can specify in the likelihood is the probability of observing the data $ \mathcal{N}_{ij}^{(q)}$, given an assignment $\pmb{z}$ of the points. The way to enforce Eq. \ref{Eq: uniformitypost} is assuming that \emph{the first neighbors of each point are preferentially points of the same manifold}. 
Consider the $i$-th row of the neighbor matrix, $\boldsymbol{\mathcal{N}}_{i}^{(q)}\equiv\{\mathcal{N}_{ij}^{(q)},j=1,\dots,N\}$. $\boldsymbol{\mathcal{N}}_{i}^{(q)}$ is a vector containing $q$ ones and $N-q$ zeros. Without any assumption, all configurations of $q$ zeros and $N-q$ ones 
are equally likely. Instead, we assume that neighbors are preferentially points from the same manifold. 
Formally, we assume that neighbors are selected forn the points of the same manifold with probability $\xi$
and from a different manifold with probability $1-\xi$, with $\xi > 1/2$. Correspondingly, we introduce a new term in the 
likelihood:
\begin{equation}
\mathcal{L}(\boldsymbol{\mathcal{N}}_{i}^{(q)}|\pmb{z})
=\frac{\xi^{n_{i}^{in}(\pmb{z})}(1-\xi)^{q-n_{i}^{in}(\pmb{z})}}{\mathcal{Z}(\xi,N_{z_{i}})},\label{eq: likelyhood neighbors}
\end{equation}
where 
\begin{equation}
n_{i}^{in}(\pmb{z})=\sum_{j}\mathcal{N}_{ij}^{(q)}\mathbb{I}_{z_{j}=z_{i}}\label{eq: n_i_in}
\end{equation}
is the number of neighbors of $i$ sampled from the same manifold, and 
\begin{equation}
q-n_{i}^{in}(\pmb{z})= \sum_{j}\mathcal{N}_{ij}^{(q)}\mathbb{I}_{z_{j}\neq z_{i}} 
\end{equation}
is the number of neighbors of $i$ sampled from a different manifold. Function $\mathcal{Z}$ is a normalization factor that depends on $\xi$:
\begin{equation}
\mathcal{Z}(\xi,N_{z_{i}})=\sum_{\{\boldsymbol{\mathcal{N}}_{i}^{(q)}\}}\xi^{n_{i}^{in}(\pmb{z})}(1-\xi)^{q-n_{i}^{in}(\pmb{z})}.
\end{equation}
and can be expressed in a compact way as 
\begin{equation}
\mathcal{Z}(\xi,N_{z_{i}})=(1-\xi)^{q}\binom{N-N_{z_{i}}}{q}{}_{2}F_{1}(-q,1-N_{z_{i}},N-N_{z_{i}}-q,\frac{\xi}{1-\xi}),
\end{equation}
where $_{2}F_{1}(a,b,c,x)$ is the hypergeometric function. The derivation of this expression and the details about the likelihood term in \eqref{eq: likelyhood neighbors} are presented below. 
By considering all points $i$, we obtain the global likelihood 
\begin{equation}
\mathcal{L}(\boldsymbol{\mathcal{N}}^{(q)}|\pmb{z}, \xi)=\prod_{i}\mathcal{L}(\boldsymbol{\mathcal{N}}_{i}^{(q)}|\pmb{z}, \xi)=\prod_{k}\frac{\xi^{n_{k}^{in}}(1-\xi)^{qN_{k}-n_{k}^{in}}}{\mathcal{Z}(\xi,N_{k})^{N_{k}}}
\end{equation}
where
\begin{equation}
n_{k}^{in}=\sum_{ij}\mathcal{N}_{ij}^{(q)}\mathbb{I}_{z_{i}=k}\mathbb{I}_{z_{j}=k}
\end{equation}
is the total number of  ``internal'' neighbors of points from manifold
$k$ and
\begin{equation}
n_{k}^{out}=\sum_{ij}\mathcal{N}_{ij}^{(q)}\mathbb{I}_{z_{i}=k}(1-\mathbb{I}_{z_{j}=k})=qN_{k}-n_{k}^{in}
\end{equation}
is the total number of ``external'' neighbors of points from $k$. Note that since $\mathcal{Z}$ depends on $i$ only through the hidden variables $\pmb{z}$ we are able to split the product into $K$ components.\\
With this additional term in the likelihood, we obtain
\begin{displaymath}
\frac{P_{post}(z_i = z_j | \mathcal{N}_{ij}^{(q)}=1, \boldsymbol{\mu},\boldsymbol{p}) }{ P_{post}(z_i = z_j | \boldsymbol{\mu},\boldsymbol{p})} = \frac{\xi}{1-\xi} > 1/2.
\end{displaymath}



\newpage

\textbf{Derivation of the neighborhood uniformity term.} 
With reference to $\boldsymbol{\mathcal{N}}_{i}^{(q)}$, without any assumption, all configurations containing $q$ zeros and $N-q$ ones  
are equally likely. It is easy to compute the number of such configurations. The problem is analogous to the problem of selecting $q$ balls from a box containing $N-1$ balls: we have to choose $q$ neighbors among $N-1$ points, point $i$ being excluded. The number of possible choices is $\binom{N-1}{q}$. Hence, all configurations of $\boldsymbol{\mathcal{N}}_{i}^{(q)}$ being equally likely we would have
\begin{displaymath}
\mathcal{L}(\boldsymbol{\mathcal{N}}_{i}^{(q)}|\pmb{z})=\binom{N-1}{q}^{-1}, \qquad \forall i.
\end{displaymath}
Instead, we assume that the neighbors of a point are preferentially points from the same manifold. 
Formally, we assume that neighbors are selected with probability $\xi$ among the $N_{z_i}$ points assigned to the same manifold of $i$, and with probability $1-\xi$ among the $N-N_{z_i}$ points assigned to a different manifold. Here $\xi > 1/2$, so that configurations with neighbors assigned to the same manifold are more likely. 
Now, the problem is analogous to the problem where we have to select $q$
balls from two boxes, a black box containing $N_b$ balls and a white one
containing $N_w$ balls. Before selecting each ball, we choose the box,
 the black one with probability $\xi$ and the white one
with probability $1-\xi$. Clearly, the probability of a choice with $n_{b}$ black and $q-n_{b}$ white 
balls is then proportional to $\xi^{n_{b}}(1-\xi)^{q-n_{b}}$.
For a given $n_b$, the number of possible choices of balls is 
\[
\binom{N_b}{n_b}\binom{N_w}{q-n_b}
\]
One can easily verify that $\sum_{n_b=0}^{q}\binom{N_b}{n_b}\binom{N_w}{q-n_b}=\binom{N_b + N_w}{q}$. The probability of a given choice is then
\[
\frac{\xi^{n_{b}}(1-\xi)^{q-n_{b}}}{\mathcal{Z}}
\]
where $\mathcal{Z}=\sum_{n_b=0}^{q}\binom{N_b}{n_b}\binom{N_w}{q-n_b}\xi^{n_b}(1-\xi)^{q-n_b}$.
By using the formula (Abramowitz and Stegun, 15.4.1)
\[
_{2}F_{1}(-m,b,c,z)=\sum_{n=0}^{m}(-)^{n}\binom{m}{n}\frac{(b)_{n}}{(c)_{n}}z^{n}
\]
where $(a)_{n}=a(a+1)\dots (a+n-1)$ is the Pochammer symbol and doing some simple algebra, $\mathcal{Z}$ can be compactly expressed as 
\begin{equation*}
\mathcal{Z}=(1-\xi)^{q}\binom{N_w}{q}{}_{2}F_{1}(-q,-N_b,N_w-q,\frac{\xi}{1-\xi}).
\end{equation*}
Replacing $N_b$ with $N_{z_{i}}-1$ (the number of points assigned to the same manifold as $i$, excluding $i$), $N_w$ with $N-N_{z_{i}}$ (the number of points assigned to a different manifold), and $n_{b}$ with $n_{i}^{in}$, we obtain the likelihood of a given configuration of $\boldsymbol{\mathcal{N}}_{i}^{(q)}$ as
\begin{equation*}
\mathcal{L}(\boldsymbol{\mathcal{N}}_{i}^{(q)}|\pmb{z},\xi)=\frac{\xi^{n_{i}^{in}(\pmb{z})}(1-\xi)^{q-n_{i}^{in}(\pmb{z})}}{(1-\xi)^{q}\binom{N-N_{z_{i}}}{q}{}_{2}F_{1}(-q,1-N_{z_{i}},N-N_{z_{i}}-q,\frac{\xi}{1-\xi})}.\end{equation*}


\newpage

\begin{figure}[b!]
\subfigure[Two Manifolds (linear)] {\includegraphics[width=0.5\columnwidth]{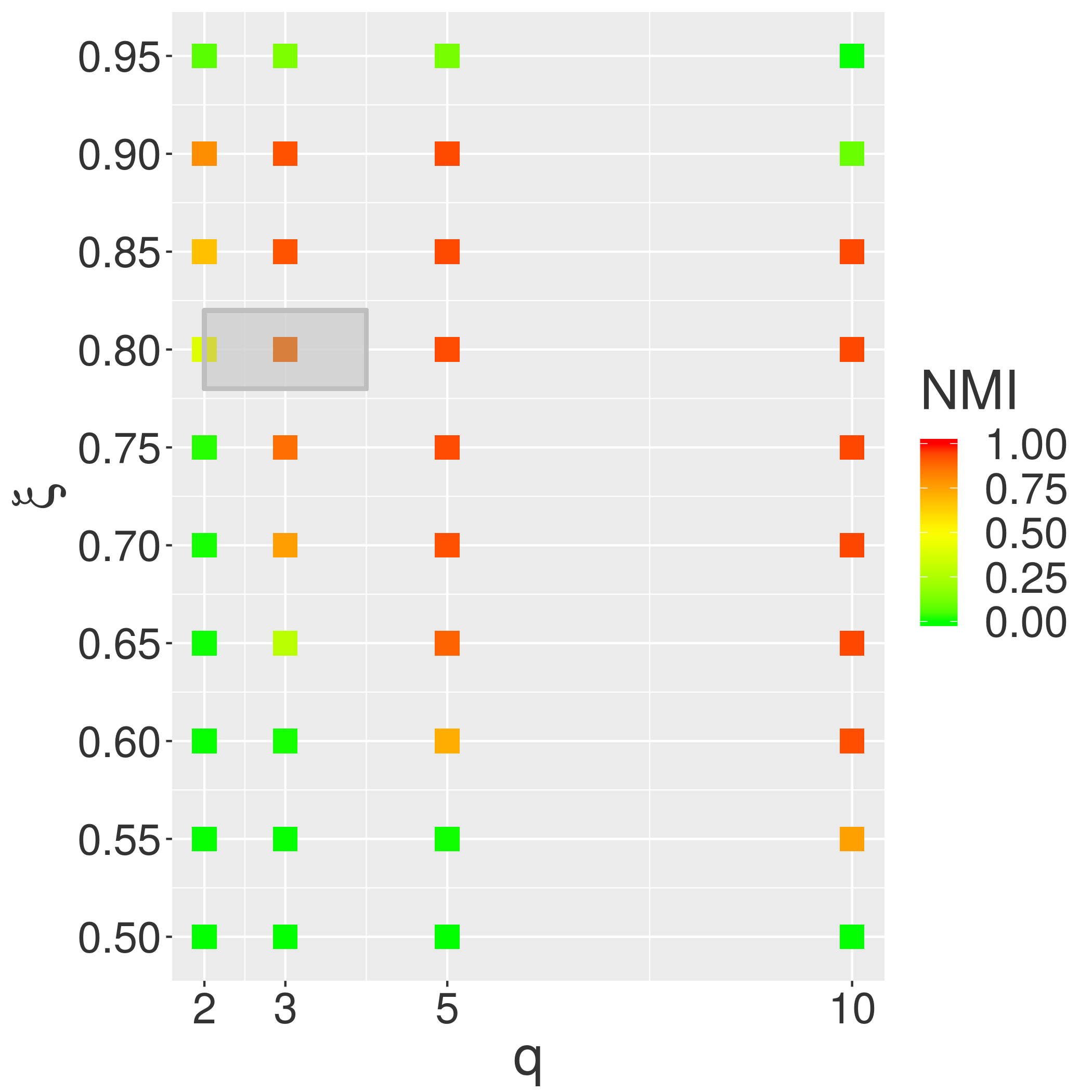}}
\subfigure[Five Manifolds (linear)] {\includegraphics[width=0.5\columnwidth]{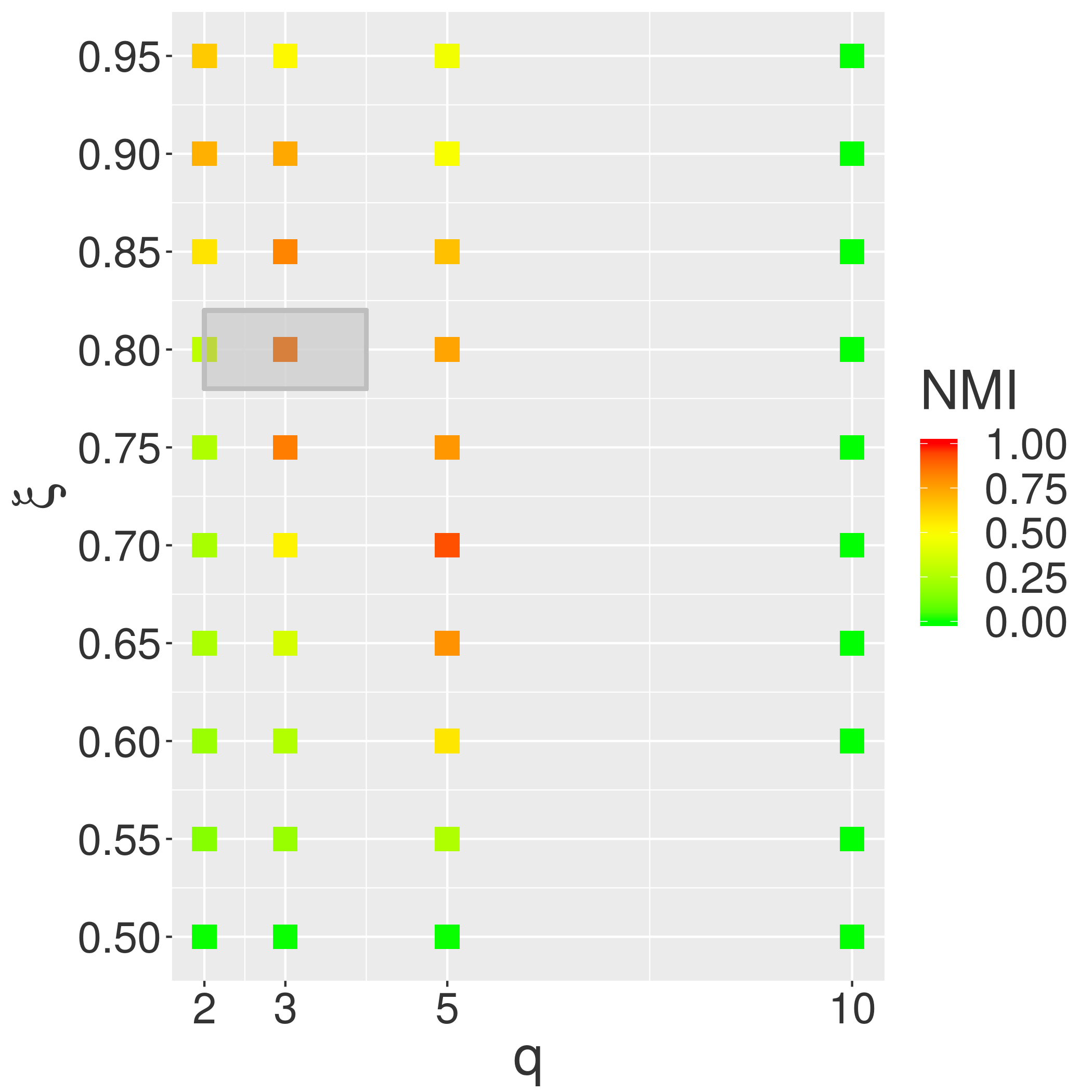}}
\subfigure[Five Manifolds (curved)] {\includegraphics[width=0.5\columnwidth]{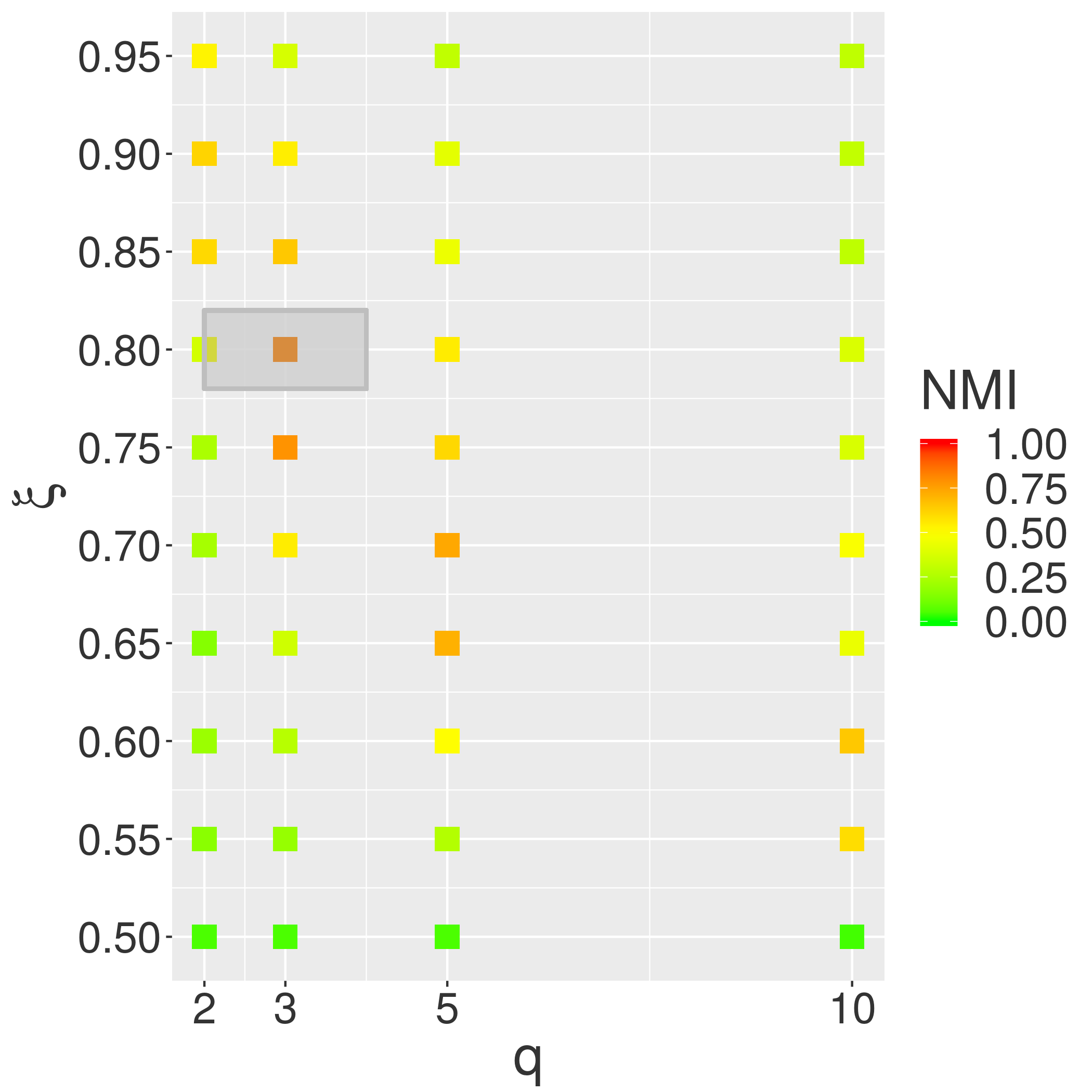}} 
\caption{\textbf{Choice of free parameters}. 
We computed the NMI between the estimated and true assignment of points for all the artificial datasets used for validation, for different valus of $q$ and $\xi$.  a) two manifolds of dimension $4$ and $5$, with a Gaussian density of points b) five manifolds of dimension 
$1,2,4,5,9$, with a Gaussian density of points c) five manifolds of dimension $1, 2, 4, 5, 9$, with a Gaussian density of points, embedded on a 1-dimensional circle, a 2-dimensional torus, a 4-dimensional Swiss roll, a 5-dimensional sphere and a 9-dimensional sphere. In all cases, we performed $10^5$ iterations of the Gibbs sampling, and repeated the sampling $M=10$ times starting from different random configurations of the parameters. We considered the sampling with the highest maximum average of the log-posterior value}
\label{Fig: parameters}
\end{figure}

\textbf{Choice of the free parameters}. In order to find a good configuration for the parameters $(q, \xi)$,
we perform tests with several values of $q \in \{2,3,\dots,10\} $ and $\xi \in [0.5,1)$ for all the artificial datasets used for validation: i) two manifolds of dimension $4$ and $5$, with a Gaussian density of points; ii) five manifolds of dimension 
$1,2,4,5,9$, with a Gaussian density of points; iii) five manifolds of dimension $1, 2, 4, 5, 9$, with a Gaussian density of points, embedded on a 1-dimensional circle, a 2-dimensional torus, a 4-dimensional Swiss roll, a 5-dimensional sphere and a 9-dimensional sphere. \\
The crucial figure of merit to assess the performance of the method is the normalized mutual information (NMI)  between the estimated and the true assignment of $\pmb{z}$, which measures the quality of the assignment of points to manifolds. Indeed, once the manifolds are correctly separated, the problem is essentially reduced to a dimension estimation within the single manifolds (which is successfully solved by TWO-NN). In Fig.~\ref{Fig: parameters}  we show the NMI as a function of $(q, \xi)$. In all cases, we performed $10^5$ iterations of the Gibbs sampling and repeated the sampling $M=10$ times starting from different random configurations of the parameters. We kept the sampling with the highest maximum average of the log-posterior value. \\
For all values of $q$, the MI first increases and then decreases with $\xi$. This can be expected based on the following considerations. When $\xi$ is close to $0.5$, as we discussed above, the method cannot discriminate between different manifolds. When $\xi$ is increased, the posterior distribution starts to prefer configurations that approximately satisfy the neighborhood homogeneity constraint. For sufficiently high $\xi$, the posterior distribution is sharply peaked at the configuration that optimally satisfies this constraint; correspondingly, if the Gibbs sampler can explore the parameter space exhaustively, it will eventually find this peaked region and remain trapped there. Hence, the NMI achieves average values close to 1. However, for $\xi$ close to $1$, the posterior distribution is very likely to also have pronounced local maxima. Therefore, depending on the initial configuration, the sampler may remain trapped in one of them. 
Hence, one can observe a drop in the NMI. 
In general, these sampling issues can be worsened when $q$ is increased since the local maxima become more and more pronounced. In principle, this problem may be dealt with by resorting to well established enhanced sampling techniques.
For simplicity, in the present work, we prefer to verify that there is a region of the parameter space where the results appear optimal and restrict to these regions for subsequent analyses. \\
In general, $q=2$ yields poor results ($NMI\lesssim 0.4$) in all cases. This means that is general $q=2$ is too low to effectively enforce the uniformity constraint. Analogously, $q=10$ too high yields poor results ($NMI\lesssim 0.5$), except in the simple case of two manifolds. There are two independent reasons for this behavior. First, $q=10$ can lead to sharp peaks in the posterior, and hence to sampling issues; second, $q=10$ gives issues in the case of intersecting manifolds, as it enforces uniformity on too large a scale. Good results ($NMI\gtrsim 0.75$) are found for $q=2, 0.75 \leq \xi \leq ,0.80,0.85$ and for $q=5, 0.65 \leq \xi \leq 0.70$. Based on these findings, we identify the optimal ``working point" of the method at $q=3, \xi=0.8$, which yields $NMI > 0.80$) for all data sets considered. 
\clearpage

\newpage

\textbf{Convergence time as a function of sample size}. 

\begin{figure}[htb]
    \centering
    \includegraphics[width=0.6\textwidth]{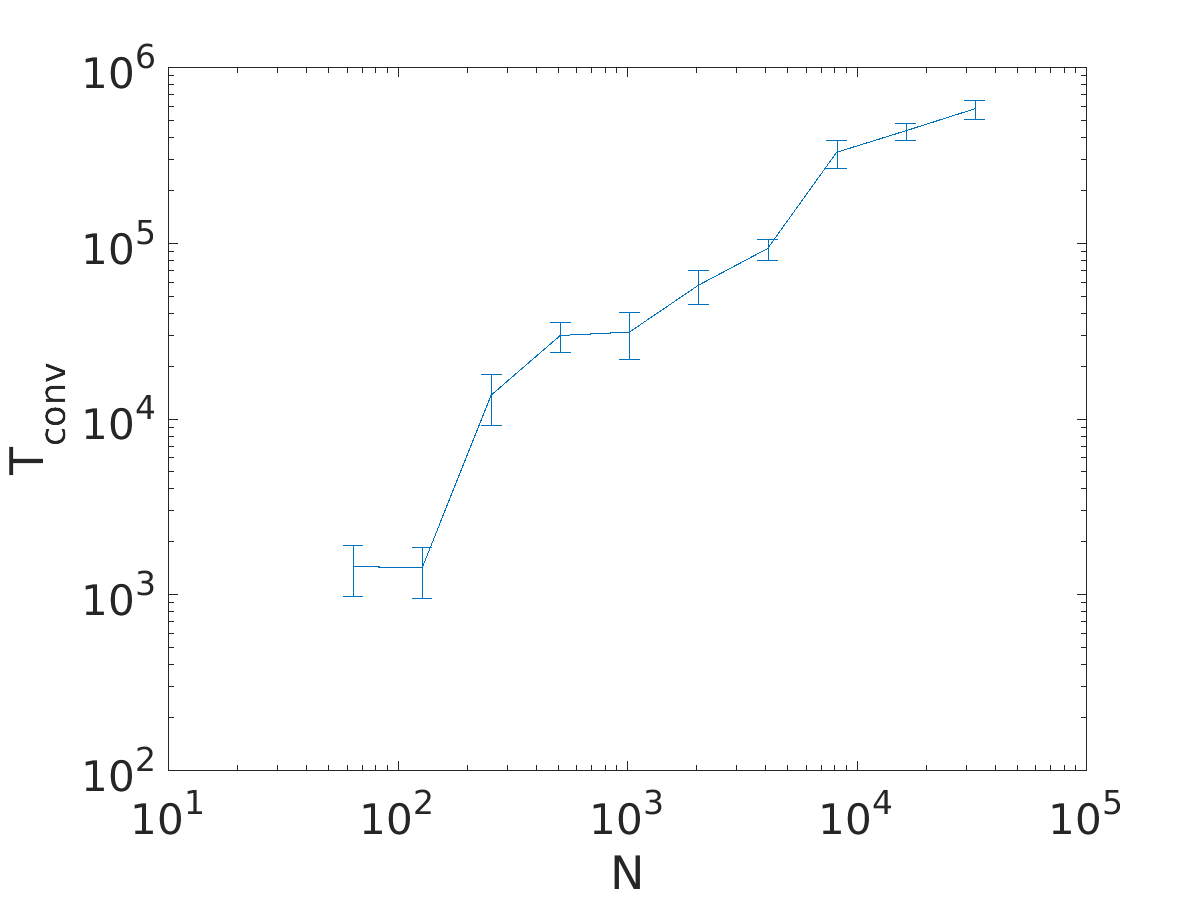}
    \caption{Convergence time $T_{conv}$ of the Gibbs sampling for $N$ points sampled from two manifolds of dimension $d_1=9$ and $d_2=4$. We repeated the estimation of $T_{conv}$ for $M=20$ independent samples. The lines connect the average value over the $M$ samples. Standard error bars are reported as well. The scaling of $T_{conv}$ with $N$ is approximately linear.}
    \label{fig:tconv}
\end{figure}

In order to test the speed of convergence of the Gibbs sampling as a function of the number of points $N$, we have generated random data sampled from two manifolds of dimension $d_1=9$ and $d_2=4$. As in Results, $N/2$ points are sampled from a multivariate Gaussian with  a variance matrix given by $1/d_i$, $i=1,2$, times the identity matrix of proper dimension. The two manifolds are embedded in a space with dimension corresponding to the higher dimension $d_2$, with their centers at a distance of $0.5$ standard deviations. We consider different sample sizes $N=2^i, \{i=6,15 \}$. 
In order to test convergence, we use the criterion in~\cite{brooks1998quantitative}. Given independent samples $Y_t$ (with $t=1,\dots,T$ and $T$ the total number of samples), and given a function $f_t=f(Y_t)$, convergence can be verified from the behavior of the cumulative sums
\begin{equation}
    S_t =  \sum_{\tau=t_0+1}^t (f_\tau - \mu) 
\end{equation}
where $ \mu = \frac{1}{T-t_0} \sum_{\tau=t_0+1}^T f_\tau $, and $t_0 \geq 1 $ is any time, usually chosen as higher to the putative burn-in time. At convergence, the behavior of $F_t$ is highly non-smooth; In particular, consider the indicator variable
\begin{equation}
e_t =   \left\{\begin{matrix}
1 \mbox{ if } S_t>S_{t-1} \mbox{ and } S_t < S_{t-1}  \\
1 \mbox{ if } S_t<S_{t-1} \mbox{ and } S_t > S_{t-1}  \\ 
0 \mbox{ otherwise} 
\end{matrix}\right. 
\end{equation}
which indicates whether $S_t$ has a local maximum/minimum at $t$. At convergence, approximately half of the points should be local minima or maxima. Formally, $e_t$ behaves as a Bernoulli variable $\mathcal{B}(1/2)$, and consequently
\begin{equation}
    E_t =  \sum_{\tau=t_0+1}^{t-1} e_\tau
    \label{Eq: et}
\end{equation}
behaves as a binomial variable $B(t-t_0-1,1/2)$. Therefore, one should have
\begin{equation}
 1/2 - Z_{\alpha/2}/\sqrt{t-t_0-1}  \leq  E_t \leq 1/2 + Z_{\alpha/2}/\sqrt{t-t_0-1}
 \label{Eq: convbounds}
\end{equation}
for at least $(1-\alpha)\%$ of the time (for large $t$, the binomial will be approximately normal). In other words, when convergence is reached, $E_t$ should fluctuate around $1/2$ with a variance compatible with that of a binomial variable. \\
To adopt this test to the data, we studied the point-wise dimension estimates $d_{1t}$ and $d_{2t}$. We subsampled the $d_{it}$ at $t=1,\theta,2\theta,\dots$ with $\theta=10$ to ensure independence of the samples. Let $D_{1t}$ and $D_{2t}$ be the resulting samples. We used $f_{1\tau}=d_{1t}$ and $f_{2\tau}=d_{2t}$ and first verified whether the corresponding $E_{1t}$, $E_{2t}$  (Eq.~\ref{Eq: et}) converged to $1/2$. While an excessive variance around $1/2$ is a sign of residual non-stationarity, failure of $E_t$ to approach the value $1/2$ for increasing $t$ is a likely sign of the non-independence of the samples used~\cite{brooks1998quantitative}. Therefore, if $E_{it}$ did not approach to $1/2$, we increased the subsampling at $\theta=20,\theta=50,\theta=100,\theta=200,\theta=500$ until $\langle E_{it} \rangle \simeq 1/2$ for large $t$. We divided the $D_{1t}$ and $D_{2t}$ in non-overlapping windows of $20$ successive points and checked whether $E_{it}$ was within the bounds (\ref{Eq: convbounds}) with $\alpha=0.05$ for 95\% of the times (at 19 times out of 20) within each window. We then defined as convergence time $T_{conv}$ the time corresponding to the first window where 
both $E_{1t}$, $E_{2t}$ were satisfying the criterion. \\
For each dataset size $N$, we repeated the assessment of $T_{conv}$ for $M=20$ independent Gibbs samplings.  
The results are shown in Fig. \ref{fig:tconv}: one can see that the convergence time increases roughly linearly with data size $N$. 

\newpage

\textbf{Restricting the analysis to points with no common first and second neighbor}.

\begin{table}[h!]
    \centering
    \small
    \begin{tabular}{| C{5cm} |  C{1.5cm} | C{1.5cm} | C{5cm} | C{1.5cm} |  }
    \hline
        \textbf{Dataset} &  $\mathbf{N_0}$ & $\mathbf{N}$ & $\mathbf{d_i}$ & \textbf{NMI}   \\
            \hline
      2 Gaussians, $d_1=5,d_2=4$ & 2000  & 357 & $d_1=5.5,\ d_2=3.9$ & 0.95 \\
    \hline
          2 Gaussians, $d_1=6,d_2=4$ & 2000  & 360 & $d_1=6.2,\ d_2=4.1$ & 0.93 \\
              \hline
          2 Gaussians, $d_1=7,d_2=4$ & 2000  & 355 & $d_1=6.5,\ d_2=4.3$ & 0.93 \\
              \hline
          2 Gaussians, $d_1=8,d_2=4$ & 2000  & 365 & $d_1=9.1,\ d_2=3.9$ & 0.96 \\
           \hline
          5 Gaussians (linear), $d_1=1,d_2=2,d_3=4,d_4=5,d_5=9$  & 5000  & 895 & $d_1=0.9,d_2=1.7,d_3=3.2,d_4=4.4,d_5=9.4$ & 0.88  \\
          \hline
          5 Gaussians (curved), $d_1=1,d_2=2,d_3=4,d_4=5,d_5=9$  & 5000  & 894 & $d_1=0.9,d_2=1.7,d_3=3.2,d_4=4.4,d_5=9.4$ & 0.61 \\
        \hline
    \end{tabular}
    \caption{\textbf{Independent points.} We repeated the analysius of artifical data sets, using only indpedendent points with non-overlapping first and second neighbor For each scenario we show the number of points in the original data set ($N_0$), the number of points upon restriction $N$, the estimated value of the $d_k$, and NMI between the assignment and the ground truth. All results were obtained with $\xi=0.8,q=3$, repeating the sampling $M=10$ times, and considering the sampling with the highest average log-posterior.  }
    \label{tab:ind}
\end{table}

\newpage

\textbf{Compustat variables used}. 

\begin{table*}[h!]
\begin{tiny}
\begin{tabular}{|l|l|l|l|l|l|} \hline
 \textbf{\#} & \textbf{Variable} & \textbf{\#} & \textbf{Variable} & \textbf{\#} & \textbf{Variable} \\ \hline
1 & Acquisitions & 12 & Liabilities - Total & 23 & Interest and Related Expense - Total\\ \hline	
2 & Assets - Total & 13 & Net Income (Loss) & 24 & Goodwill \\ \hline	
3 & Capital Expenditures & 14 & Operating Income Before Depreciation & 25 & Intangible Assets - Total \\ \hline
4 & Cash & 15 & Property Plant and Equipment - Total (Net) & 26 & Pretax Income \\ \hline
5 & Common Shares Outstanding & 16 & Purchase of Common and Preferred Stock & 27 & Pretax Income - Foreign \\ \hline
6 & Common/Ordinary Shareholders & 17 & Sales/Turnover (Net) & 28 & Investment and Advances - Equity \\ \hline
7 & Debt in Current Liabilities - Total & 18 & Stockholders Equity - Parent & 29 & Investment and Advances - Other \\ \hline
8 & Long-Term Debt - Total & 19 & Income Taxes Paid & 30 & Increase in Investments \\ \hline
9 & Cash Dividends (Cash Flow) & 20 & Research and Development Expense & 31 & Sale of Investments \\ \hline
10 & Earnings Before Interest and Taxes & 21 & Price Close - Annual - Fiscal & \ & \ \\ \hline
11 & Employees & 22 & Preferred/Preference Stock (Capital) - Total  & \ & \ \\ \hline
\end{tabular}
\end{tiny}
\caption{\textbf{Compustat variables used}}
\end{table*}

\end{document}